\documentclass[10pt,journal,compsoc,hidelinks]{IEEEtran}
\ifCLASSOPTIONcompsoc
  \usepackage[nocompress]{cite}
\else
  \usepackage{cite}
\fi
\usepackage{amsmath,amssymb}
\usepackage{url}
\usepackage{graphicx}
\usepackage{multirow,multicol}
\usepackage{orcidlink}
\usepackage{pifont}
\newcommand{\cmark}{\ding{51}}
\newcommand{\xmark}{\ding{55}}
\newcommand{\bclr}[1]{{\color{black}#1}}

\begin{document}
%

\title{Hybrid Open-set Segmentation \\ with Synthetic Negative Data}

\author{Matej~Grcić\orcidlink{0000-0002-8379-6686
},
        Siniša~Šegvić\orcidlink{0000-0001-7378-0536}
\thanks{M. Grcić and S. Šegvić are with University of Zagreb Faculty of Electrical Engineering and Computing, Unska 3, 10000 Zagreb, Croatia\protect\\
E-mail: \{matej.grcic,sinisa.segvic\}@fer.hr}
\thanks{Manuscript received January 19, 2023.}}

%
%

\markboth{Journal of \LaTeX\ Class Files,~Vol.~14, No.~8, August~2015}%
{Shell \MakeLowercase{\textit{et al.}}: Bare Demo of IEEEtran.cls for Computer Society Journals}
%



\IEEEtitleabstractindextext{%
\begin{abstract}
Open-set segmentation can be conceived
by complementing closed-set classification
with anomaly detection.
Many of the existing dense anomaly detectors operate
 through generative modelling
of regular data or
by discriminating with respect
to negative data.
These two approaches optimize different objectives
and therefore exhibit different failure modes.
Consequently, we propose a novel anomaly score
that fuses generative and discriminative cues.
Our score can be implemented
by upgrading any closed-set segmentation model
with dense estimates
of dataset posterior and unnormalized data likelihood.
The resulting dense hybrid open-set models
require negative training images
that can be sampled from
an auxiliary negative dataset,
from a jointly trained generative model,
or from a mixture of both sources.
We evaluate our contributions
on benchmarks for dense anomaly detection
and open-set segmentation.
The experiments reveal strong
open-set performance in spite of negligible computational overhead.
\end{abstract}

\begin{IEEEkeywords}
Open-set segmentation, Open-set recognition, Out-of-distribution detection, Anomaly detection, Semantic segmentation, Synthetic data, Normalizing flows, Hybrid models
\end{IEEEkeywords}}

\maketitle
\IEEEdisplaynontitleabstractindextext

%
\IEEEpeerreviewmaketitle

\IEEEraisesectionheading{\section{Introduction}\label{sec:introduction}}

%
%
%
%
\IEEEPARstart{H}{igh} accuracy, fast inference and small memory footprint of modern neural networks \cite{he16cvpr,liu21iccv} steadily expand the horizon of downstream applications.
Many exciting applications require advanced image understanding provided by semantic segmentation \cite{everingham10ijcv,farabet13pami,minaee22tpami}.
These models associate each pixel with a class from a predefined taxonomy \cite{shelhamer17pami,chen18tpami,strudel21iccv},
and can deliver reasonable performance on two megapixel images in real-time on low-power hardware \cite{you20eccv,orsic21pr,pan2022tits}.
However, standard training procedures assume closed-world setup, which may raise serious safety issues in real-world deployments \cite{blum21ijcv,gonzalez22mia}. 
For example, if a segmentation model missclassifies an unknown object (e.g. lost cargo) as road, the autonomous car may experience a serious accident \cite{pinggera16iros}.
Such hazards can be alleviated by complementing semantic segmentation with dense anomaly detection \cite{ruff21pieee,boult19aaai}.
The resulting open-set models are fitter for real applications due to ability to decline decisions in unknown scene parts \cite{bevandic22ivc,cen21iccv,kong22tpami}.

Previous approaches for dense anomaly detection assume either generative or  discriminative perspective.
Generative approaches are based on density estimation \cite{blum19iccvw,du22iclr}, adversarial learning \cite{kong22tpami} or image resynthesis \cite{lis19iccv,biase21cvpr,vojir21iccv}.
Discriminative approaches rely on classification confidence \cite{devries18arxiv}, dataset posterior \cite{bevandic22ivc} or Bayesian inference \cite{kendall17nips}.
However, the two perspectives exhibit different failure modes.
Generative anomaly detectors inaccurately disperse 
the probability volume \cite{nalisnick19iclr,serra20iclr,lucas19nips,zhang21icml} or face ill-posedness of image resynthesis \cite{lis19iccv,biase21cvpr}.
On the other hand, discriminative anomaly detectors produce confident predictions at arbitrary distances from the training data.
This shortcoming can be mitigated 
by training on negative content
collected from some general-purpose auxiliary dataset \cite{bevandic22ivc,biase21cvpr,chan21iccv}.
However, such training may involve an overlap between the negatives and validation anomalies.
Hence, the evaluation may lead to over-optimistic estimates and fatal production failures.
We combine the two perspectives by complementing closed-set segmentation with unnormalized dense data likelihood $\hat{p}(\mathbf{x})$ and dense dataset posterior $P(d_\mathrm{in}|\mathbf{x})$.
Fusion of the latter two predictions yields an effective yet efficient dense anomaly detector which we refer to as DenseHybrid.
Both components of our anomaly detector require training with negative content \cite{hendrycks19iclr,lee18iclr,bevandic22ivc,biase21cvpr}.
Nevertheless, our experiments evaluate performance with and without real negative data.

\begin{figure*}
    \centering
    \includegraphics[width=\linewidth]{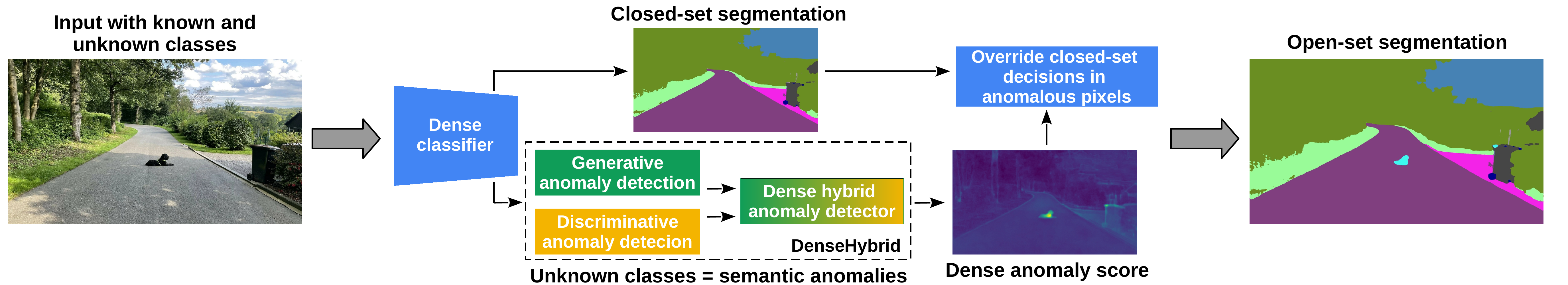}
    \caption{\bclr{
    Open-set segmentation simultaneously classifies known scene parts and identifies unknown classes (highlighted in cyan). 
    Our approach exploits the fact that unknown classes are semantic anomalies \cite{ruff21pieee}. Hence, we construct a dense hybrid anomaly detector and use it to override closed-set decisions in anomalous pixels.
Our hybrid anomaly score identifies pixels as unknown visual concepts
by efficient ensembling of generative and discriminative predictions.
    }}
    \label{fig:ood_detector2}
\end{figure*}

This paper extends our preliminary conference report \cite{grcic22eccv} by allowing our dense hybrid models to train without real negative data.
We achieve that by generating synthetic negative samples with a jointly trained normalizing flow.
Our new experiments explore
open-set training without real negative data,
validate our unnormalized density
against a generative baseline,
and compare our synthetic negatives
with alternative approaches \cite{lee18iclr,besnier21iccv,rombach22cvpr}.
Moreover, we test our method
on two novel setups, Pascal-COCO and COCO20/80, 
with diverse unknown unknowns. 

Our consolidated work brings forth the following contributions. 
First, we propose the first hybrid anomaly detector that allows pixel-level predictions, translational equivariance,
and end-to-end coupling with semantic posterior.
Our detector achieves these properties
by combining unnormalized density
with discriminative dataset posterior.
Note that translational equivariance is a non-optional ingredient in most dense prediction setups \cite{shelhamer17pami}.
If our unnormalized density estimates were recovered by evaluating a crop-wide generative model in a sliding window, the computational cost would greatly exceed the complexity of closed-set semantic segmentation and thus preclude most practical applications.
Second, we upgrade our approach by extending it to learn only on inlier images.
This configuration leverages synthetic negative data 
produced by a jointly trained generative model.
Third, we propose open-mIoU as a novel performance metric for open-set segmentation in safety-critical applications. 
The main strength of the novel metric is the exact quantification of the gap between closed-set and open-set evaluation.
Fourth, our hybrid anomaly detector can be easily attached to any closed-set segmentation approach that optimizes pixel-level cross-entropy.
The resulting open-set segmentation models deliver competitive performance 
on standard benchmarks 
with and without training on real negative data.

\section{Related Work}

Our review considers anomaly detection (Sec.\ \ref{sec:rw_iwood} and Sec.\ \ref{sec:rw_pwood}), open-set recognition (Sec.\ \ref{sec:rw_osr}), training on synthetic data (Sec.\ \ref{sec:rw_syn}), and advances towards open worlds (Sec.\ \ref{sec:rw_byn}).

\subsection{Image-wide Anomaly Detection}
\label{sec:rw_iwood}
Detecting samples which deviate from the generative process of the training data is a decades-old problem \cite{hawkins80book}.
This task calls for the same methods as novelty detection and out-of-distribution (OOD) detection \cite{hendrycks17iclr,ruff21pieee}.
Early image-wide approaches
utilize max-softmax probability \cite{hendrycks17iclr}, input perturbations \cite{liang18iclr}, ensembling \cite{lakshminarayanan17nips} or Bayesian uncertainty \cite{kendall17nips}.
More encouraging performance has been attained through discriminative training against real negative data \cite{dhamija18nips,hendrycks19iclr,bevandic22ivc,liu20neurips}, adversarial attacks \cite{besnier21iccv} or samples from generative models \cite{lee18iclr,neal18eccv,grcic21visapp,kong22tpami}.

Another line of work detects anomalies by estimating the data likelihood.
Surprisingly, this research reveals that anomalies may give rise to higher likelihood than inliers \cite{nalisnick19iclr,serra20iclr,zhang21icml}.
Generative models can mitigate this problem by sharing features
with the primary discriminative model \cite{zhang20eccv}
and training on negative data \cite{hendrycks19iclr}.  

\subsection{Pixel-wise Anomaly Detection}
\label{sec:rw_pwood}

Image-wide anomaly detectors can be adapted for dense prediction with variable success.
Some
image-wide approaches
are not applicable in dense prediction setups \cite{zhang20eccv},
while others do not perform well \cite{hendrycks17iclr} or involve excessive computational complexity \cite{liang18iclr,lakshminarayanan17nips}.
On the other hand, discriminative training with negative data \cite{hendrycks19iclr,dhamija18nips,bevandic19gcpr} is easily ported to dense prediction.
Hence, several dense anomaly detectors are trained on mixed-content images obtained by pasting negative content (e.g.\ ImageNet1k, COCO, ADE20k) over regular training images \cite{bevandic22ivc,chan21iccv,biase21cvpr}.
Dataset posterior can be recovered by sharing features with the standard segmentation head \cite{bevandic22ivc}.

Anomalies can also be recognized in a pre-trained feature space \cite{blum19iccvw}.
However, this approach is sensitive to feature collapse \cite{lucas19nips}.
Orthogonally, anomaly detectors can be implemented according to learned dissimilarity between the input and resynthesized images \cite{lis19iccv,xia20eccv,biase21cvpr,vojir21iccv}.
The resynthesis can be performed by a generative model conditioned on the predicted labels.
This approach is suitable only for uniform backgrounds such as roads \cite{lis19iccv} and offline applications due to being ill-posed.
Furthermore, it involves a large computational overhead.
Some approaches consider applications without the primary task \cite{zavrtanik21pr},
however, these are less relevant for our open-set setups.

Different than all previous work, we propose the first hybrid anomaly detector for dense prediction models.
In comparison with previous approaches that build on dataset posterior \cite{bevandic22ivc,hendrycks19iclr,dhamija18nips}, our method introduces synergy with unnormalized likelihood evaluation. 
In comparison with approaches that recover dense likelihood \cite{blum21ijcv}, our method introduces joint hybrid end-to-end training and efficient joint inference together with standard semantic segmentation.
Our method is related to joint energy-based models \cite{grathwohl20iclr,zhao23pami}, since we also reinterpret logits as unnormalized joint likelihood. 
However, previous energy-based approaches have to backprop through the intractable normalization constant and are therefore unsuitable for large resolutions and dense prediction. 
Our method avoids model sampling by recovering unnormalized likelihood and training on negative data.
Concurrent works \cite{tian22eccv,liang22neurips} consider 
alternative formulations of our generative component.

\subsection{Open-set Recognition}
\label{sec:rw_osr}
Open-set recognition assumes presence of test examples that transcend the training taxonomy.
Such examples are also known as semantic anomalies \cite{ruff21pieee}.
During inference, the model has to recognize semantic anomalies
and withhold (or reject) the decision \cite{scheirer12tpami}.
The rejection mechanism can be implemented by restricting the shape of the decision boundary \cite{scheirer14tpami,bendale16cvpr}.
This can be carried out by thresholding  the distance from learned class centers in the embedding space \cite{scheirer14tpami,cen21iccv}.
However, this approach can not deal  
with outliers that get embedded 
nearby the class centers.
Alternatively, the rejection can be formulated by complementing the classifier with a semantic anomaly detector \cite{boult19aaai,hendrycks17iclr,liang18iclr}.
Open-set performance can be further improved by supplying more capacity \cite{vaze22iclr,chen22tpami}.
More details about open-set approaches can be found in the recent review \cite{geng21tpami}.

Most open-set approaches quantify performance by separate evaluation of closed-set recognition and anomaly detection
\cite{hendrycks17iclr,zendel18eccv,blum21ijcv,chan21neuripsd}.
However, such practice does not reveal degradation of discriminative predictions due to errors in anomaly detection \cite{sokolova09ipm,scherreik16taes}.
This is especially pertinent to dense prediction where we can observe inlier and outlier pixels in the same image.  
Recent work proposes a solution for the related problem
of semantic segmentation in adverse conditions \cite{sakaridis22pami}.
Their uncertainty-aware UIoU metric takes into account prediction confidence as measured by the probability of the winning class.
However, UIoU assumes that each pixel belongs to one of the known classes, which makes it inapplicable for open-set setups.
Different than all previous work, our open-IoU metric specializes for open-set segmentation in the presence of outliers.
It takes into account both false positive semantic predictions at outliers as well as false negative semantic predictions due to false positive anomaly detection.
Furthermore, the difference between mIoU and open-mIoU reveals the performance gap caused by outliers.

\subsection{Synthetic Negative Data in Open-set Recognition}
\label{sec:rw_syn}

Recent seminal approaches train open-set recognition models on synthetic negative data produced by a jointly trained generative adversarial network \cite{neal18eccv,lee18iclr}.
The GAN is trained to generate inlier data that give rise to low recognition scores for each known class \cite{lee18iclr}.
However, GANs are biased towards limited distribution coverage \cite{lucas19nips}. 
Consequently, they are unlikely to span the whole space of possible outliers.
Thus, more promising results were achieved by mixing real and synthetic negative samples \cite{kong22tpami}.

Distributional coverage can be improved
by replacing GANs with generative models 
that optimize likelihood \cite{lucas19nips}. 
Our context calls for efficient approaches with fast sampling since joint training requires sample generation on the fly. 
This puts at a disadvantage autoregressive PixelCNN and energy-based models.
Normalizing flows are a great candidate for this role due to fast training and capability to quickly generate
samples at different resolutions \cite{grcic21visapp}.
Instead of targeting negative data, a generative model can also target
negative features \cite{kong22tpami}.
This can be done by modelling inlier features 
and sampling synthetic anomalies from low-likelihood regions of feature space \cite{du22iclr,kumar23cvpr}.
Synthetic negative data can also be crafted by leveraging
adversarial perturbations \cite{besnier21iccv}.

\subsection{Beyond Open-set Recognition}
\label{sec:rw_byn}
Open-world recognition can be formulated
by clustering anomalous images or representations 
into new semantic classes.
This can be done in incremental \cite{michieli21cviu,uhlemeyer22uai} or zero/one/few-shot \cite{fu20tpami} setting.
However, these approaches are still unable to compete with supervised learning on standard datasets.
We direct the reader to \cite{xian19tpami} for an exhaustive analysis of pros and cons of low-shot learning. 

\begin{figure*}[ht]
    \centering
    \includegraphics[width=0.95\linewidth]{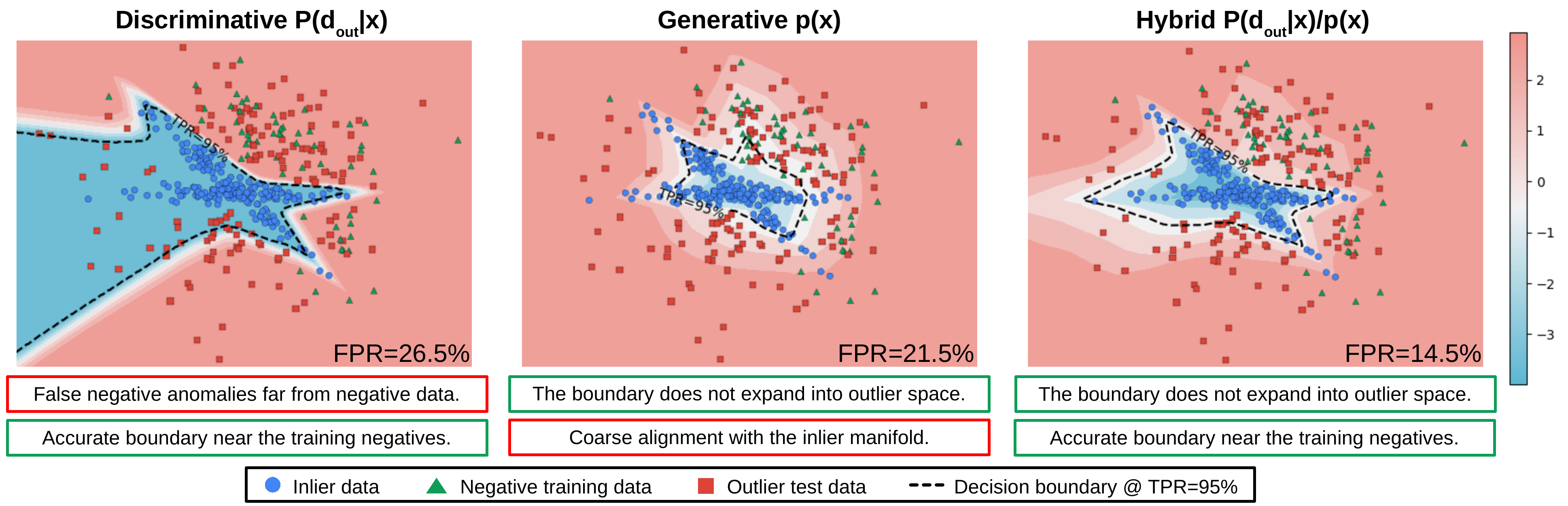}
    \caption{
Three anomaly detection approaches on a toy problem.
Inliers, train negatives and test anomalies are shown
as blue, green and red points (details in the Appendix).
The background heatmaps designate the three anomaly scores with higher values in red. 
The discriminative anomaly score (left) is susceptible to false negative responses since the negative training dataset is finite and cannot cover all modes of test anomalies. 
The generative anomaly score (middle) errs along the border of the inlier manifold
due to over-generalization
\cite{nalisnick19iclr,lucas19nips},
but is unlikely to commit errors far from the inlier manifold.
Our hybrid approach prevails by ensembling
 discriminative and generative cues.
}
    \label{fig:osr_poc}
\end{figure*}


\section{Hybrid  Open-set Segmentation}


We express open-set segmentation
by overriding closed-set predictions
with a novel dense hybrid anomaly score (Sec.\ \ref{sec:ratio_train}).
Our score incorporates
discriminative and generative cues
built atop shared semantic features (Sec.\ \ref{sec:prob_view})
that enable efficient open-set inference (Sec.\ \ref{sec:hybrid_oss}).


We represent input images $\mathbf{x} \in \mathcal{X}$ with a random variable $\underline{\mathbf{x}}$.
Categoric random variable $\underline{y}^{ij}$ denotes  label at the location ($i,j$).
Binary random variable $\underline{d}^{ij}$ models whether the given pixel is an inlier or as outlier. 
We write $d_\mathrm{in}^{ij}$ for inliers and $d_\mathrm{out}^{ij}$ for outliers.
Finally, we denote a realization of a random variable 
by omitting the underline
and often omit spatial locations for brevity.
Thus, $P(y|\mathbf{x})$ and $p(\mathbf{x})$ are shortcuts for $P(\underline{y}^{ij}=y^{ij}|\underline{\mathbf{x}}=\mathbf{x})$  and $p(\underline{\mathbf{x}}^{ij} = \mathbf{x}^{ij})$. 

\subsection{Hybrid Anomaly Detection
}
\label{sec:ratio_train}
Figure \ref{fig:osr_poc} presents three 
anomaly detection approaches
on a 2D toy problem.
The discriminative approach 
models the dataset posterior $P(d_\mathrm{in}|\mathbf{x})$. 
It often fails far from the inliers
since a finite negative training dataset
cannot cover all modes of the test anomalies. 
The generative approach models 
the data likelihood $p(\mathbf{x})$.
It often errs 
along the boundary of the inlier manifold 
due to over-generalization \cite{nalisnick19iclr,lucas19nips}, 
but does not expand into the open space.
We ensemble these two approaches
since they tend to assume different failure modes.
Our hybrid anomaly score alleviates 
both the coarseness of the generative approach 
and the over-confidence of the discriminative approach.
This synergy favours accurate boundaries 
near the negative training data 
while reducing false negative anomalies
in the open space.

More formally, we can state a sufficient
condition for performance gain of our hybrid ensemble over each of its two components.
Let $s: \mathcal{X} \rightarrow \mathbb{R}$ be a standardized anomaly score which assigns higher values to anomalies.
We can decompose the score $s$ into correct labeling $f$ and error $\epsilon$:
\begin{equation}
    s(\mathbf{x}) = f(\mathbf{x}) + \epsilon(\mathbf{x}).
\end{equation}
Function $f: \mathcal{X} \rightarrow \{-1, +1\}$ labels anomalies with +1 and inliers with -1.
The expected squared error then equals:
\begin{equation}
    \mathcal{E}(s) = \mathbb{E}_\mathbf{x} [(s(\mathbf{x}) - f(\mathbf{x}))^2] =  \mathbb{E}_\mathbf{x} [(\epsilon(\mathbf{x}))^2].
\end{equation}
Our goal is to show conditions under which the hybrid anomaly score outperforms both of its components:
\begin{equation}
    \mathcal{E}(s_H) < \inf\{ \mathcal{E}(s_G), \mathcal{E}(s_D) \}.
    \label{eq:hybrid_ineq2}
\end{equation}
The generative anomaly score $s_G$ is a function of data likelihood.
The discriminative anomaly score $s_D$ is a function of dataset posterior.
By defining our hybrid anomaly detector as $s_H(\mathbf{x}) := \frac{1}{2} s_D(\mathbf{x}) + \frac{1}{2} s_G(\mathbf{x})$, the condition (\ref{eq:hybrid_ineq2}) becomes as follows (proof in the Appendix):
\begin{equation}
    \frac{\alpha - 3}{4} e + C_1 \rho(\epsilon_D, \epsilon_G) + C_2 < 0
    \label{eq:hybrid_effectivness_test2}
\end{equation}
Here $\rho$ is the Pearson correlation coefficient between the errors, $\alpha$ = $\frac{\sup\{ \mathcal{E}(s_G), \mathcal{E}(s_D) \}}{\inf\{ \mathcal{E}(s_G), \mathcal{E}(s_D) \}}$
denotes the error ratio of the two components, $e=\inf\{ \mathcal{E}(s_G), \mathcal{E}(s_D) \} $ denotes the smallest expected error, while $C_1$ and $C_2$ can be viewed as constants.
If the errors of the two components 
are  independent and Gaussian
($\rho=0, C_1 = 0.5$ and $C_2 = 0$), 
then our hybrid anomaly detector 
will be effective even if $\alpha<3$.
The condition (\ref{eq:hybrid_effectivness_test2}) can be
satisfied even when the two components are moderately correlated as in our experiments. 
This creates an opportunity
to build efficient dense hybrid anomaly detectors atop shared features.

\subsection{Efficient Implementation Atop Semantic Classifier}
\label{sec:prob_view}

Standard semantic segmentation can be viewed as a two-step procedure.
Given an input image $\mathbf{x}$, a deep feature extractor $f_{\theta_1}$ computes an abstract representation $\mathbf{z}$ also known as pre-logits.
Then, the computed pre-logits are projected into logits $\mathbf{s}$ and activated by softmax.
The softmax output models the class posterior $P(y|\mathbf{x})$:
\begin{equation}
\label{eq:cls-init}
    P(y|\mathbf{x}) := \mathrm{softmax}(\mathbf{s}_{y}), \;\text{where}\;  \mathbf{s} = f_{\theta_2}(\mathbf{z}), \, \mathbf{z} = f_{\theta_1}(\mathbf{x}).
\end{equation}
In practice, $f_{\theta_1}$ can be any dense feature extractor that is suitable for semantic segmentation, while $f_{\theta_2}$ is a simple projection.
We extend this framework with dense data likelihood and discriminative dataset posterior.

Dense data likelihood can be expressed atop the dense classifier $f_{\theta_2}$ by re-interpreting exponentiated logits $\mathbf{s}$ as unnormalized joint density $\hat{p}(y, \mathbf{x})$\cite{grathwohl20iclr}:
\begin{equation}
\label{eq:p_x}
    p(\mathbf{x}) = \sum_y p(y, \mathbf{x}) = \frac{1}{Z} \, \sum_y \hat{p}(y, \mathbf{x})   = \frac{1}{Z} \, \sum_y \exp{\mathbf{s}_y}.
\end{equation}
Consequently, the unnormalized likelihod corresponds to $\hat{p}(\mathbf{x}) = \sum_y \exp \mathbf{s}_y$.
$Z$ denotes the normalization constant dependent only on model parameters.
As usual, $Z$ is finite but intractable, since it requires aggregating the unnormalized distribution for all realizations of $\underline{y}$ and $\underline{\mathbf{x}}$:
 $Z = \int_\mathbf{x} \sum_y \exp{\mathbf{s}_y}$.
Throughout this work, we conveniently eschew the evaluation of $Z$ in order to enable efficient training and inference.

The standard discriminative predictions (\ref{eq:cls-init})
are consistently recovered 
as $p(y,\mathbf{x}) / p(\mathbf{x})$
according to Bayes rule \cite{grathwohl20iclr}:
\begin{equation}
   P(y|\mathbf{x}) = \frac{p(y,\mathbf{x})}{\sum_{y'} p(y',\mathbf{x})} = \frac{\exp{\mathbf{s}_y}}{\sum_{y'} \exp{\mathbf{s}_{y'}}} = \mathrm{softmax}(\mathbf{s}_y).
\end{equation}
The normalization constant $Z$ appears both in the numerator and denominator, and hence cancels out.
Reinterpretation of logits (\ref{eq:p_x}) enables convenient unnormalized per-pixel likelihood estimation 
atop pre-trained dense classifiers.
Note that adding a constant value to the logits does not affect the standard classification but affects our formulation of data likelihood.
We exploit the extra degree of freedom to 
formulate the generative anomaly score $s_G(\mathbf{x}) \propto - \ln \hat{p}(\mathbf{x})$\footnote{The same extra degree of freedom has been used to model a discriminator network in semi-supervised learning \cite{salimans16neurips}.}.

We define the dataset posterior $P(d_\mathrm{in}|\mathbf{x})$ as a non-linear transformation $g_\gamma$ of pre-logits $\mathbf{z}$ \cite{bevandic22ivc}:
\begin{equation}
    P(d_\mathrm{in}|\mathbf{x}) = 1 -  P(d_\mathrm{out}|\mathbf{x}) := \sigma(g_{\gamma}(\mathbf{z})).
\end{equation}
The discriminative anomaly scor 
$s_D(\mathbf{x}) \propto \ln P(d_\text{out}|\mathbf{x})$.
Finally, we materialize our hybrid anomaly score 
as a likelihood ratio that can also be interpreted 
as ensemble $s_H(\mathbf{x}) = s_D(\mathbf{x}) + s_G(\mathbf{x})$:
\begin{equation}
    \label{eq:sx}
    s_H(\mathbf{x}) := \ln \frac{ P(d_\mathrm{out}|\mathbf{x})}{ p(\mathbf{x})}  \cong \ln P(d_\mathrm{out}|\mathbf{x}) - \ln \hat{p}(\mathbf{x}).
\end{equation}
Our generative score can neglect $Z$ since the ranking performance \cite{hendrycks17iclr} is invariant to monotonic transformation such as taking a logarithm or adding a constant.
The detailed derivation and connection with the ensemble of the two components is in the Appendix.
Our score is well suited for dense prediction due to minimal overhead and translational equivariance.
This particular formulation equalizes the influence of the two components. 
Still, other definitions may also be effective, which is an interesting direction for future work.

\subsection{
Dense Open-set Inference}
\label{sec:hybrid_oss}

\begin{figure}[ht]
    \centering
    \includegraphics[width=\linewidth]{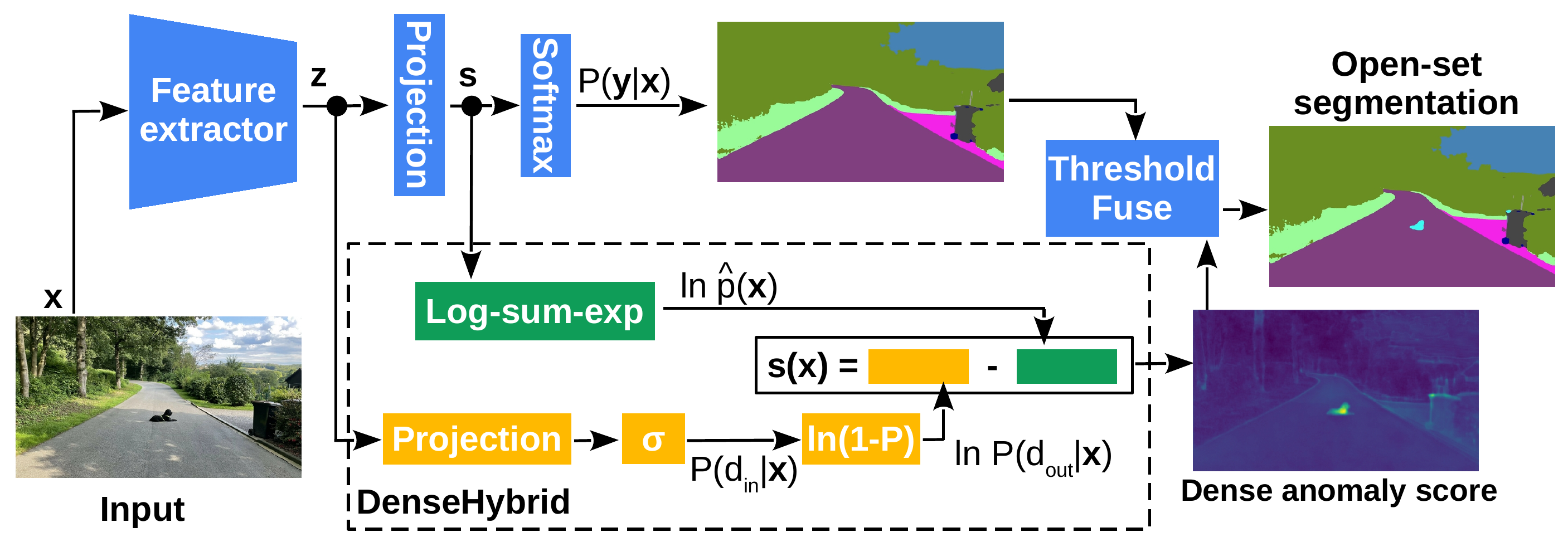}
    \caption{
    \bclr{
    Our open-set segmentation approach complements any semantic segmentation model which recovers dense logits with our hybrid anomaly detection.
  Our dense anomaly score is a log-ratio of dataset posterior and data likelihood. We implement open-set segmentation by overriding the closed-set output with thresholded anomaly score.}}
    \label{fig:ratio}
\end{figure}

The proposed hybrid anomaly detector can be combined with the closed-set output to recover open-set predictions as shown in Figure  \ref{fig:ratio}.
The input is fed to a dense feature extractor which produces pre-logits $\mathbf{z}$ and logits
$\mathbf{s}$.
We recover the closed-set posterior $P(y|\mathbf{x})$ with softmax, and the unnormalized data log-likelihood $\ln \hat{p}(\mathbf{x})$ with log-sum-exp (designated in green).
A distinct head $g$ transforms pre-logits $\mathbf{z}$ into the dataset posterior $P(d_\mathrm{in}|\mathbf{x})$ (designated in yellow).
The anomaly score $s(\mathbf{x})$ is a log ratio between dataset-posterior and density (\ref{eq:sx}).
The resulting anomaly map is thresholded and fused with the discriminative output into the final dense open-set output.
The desired behaviour of the dense hybrid open-set model is attained by fine-tuning a pre-trained classifier as we describe next.

\section{Open-set Training with DenseHybrid}

Our open-set approach complements an arbitrary closed-set segmentation model with the DenseHybrid anomaly detector.
We propose a novel training setup that eschews the intractable normalization constant by introducing negative data to the generative learning objective (Sec.\ \ref{sec:oss_train_aux}).
The same negative data is used to train the dataset posterior.
We relax dependence on real negatives by sampling a jointly trained normalizing flow (Sec.\ \ref{sec:oss_train_syn}). 

\subsection{Open-set Training with Real Negative Data}
\label{sec:oss_train_aux}

Our hybrid open-set model
requires joint fine-tuning of three dense prediction heads: 
closed-set class posterior $P(y|\mathbf{x})$,
 unnormalized data likelihood $\hat{p}(\mathbf{x})$ \cite{grathwohl20iclr}, and   
 dataset posterior $P(d_{\mathrm{in}} | \mathbf{x})$ 
\cite{bevandic22ivc}.
The corresponding training objectives
are presented in the following paragraphs.

\noindent
\textbf{Class posterior.} The closed-set class-posterior head can be trained according to the standard discriminative cross-entropy loss 
over the inlier dataset $D_\mathrm{in}$:
\begin{align}
    \nonumber
    L_{\mathrm{cls}}(\theta) &= \mathbb{E}_{\mathbf{x}, y \in D_\mathrm{in}}[- \ln P(y|\mathbf{x})] \\ &= \mathbb{E}_{\mathbf{x}, y \in D_\mathrm{in}}[ - \mathbf{s}_{y}] + \mathbb{E}_{\mathbf{x}, y \in D_\mathrm{in}}[\underset{y'}{\text{LSE}}(\mathbf{s}_{y'})].
    \label{eq:cls}
\end{align}
As before, $\mathbf{s}$ are logits computed by $f_\theta$, while LSE stands for log-sum-exp where the sum iterates over classes.

\noindent
\textbf{Data likelihood.}
Training unnormalized likelihood can be a daunting task since backpropagation through $p(\mathbf{x})$ involves intractable integration over all possible images \cite{du19neurips,song19neurips}.
Previous MCMC-based solutions \cite{grathwohl20iclr} are not feasible in our setup due to high-resolution inputs and dense prediction.
We eschew the normalization constant by optimizing the likelihood both in inlier and outlier pixels:
\begin{align}
\nonumber
L_{\mathbf{x}}(\theta) &= \mathbb{E}_{\mathbf{x} \in D_\mathrm{in}}[- \ln p(\mathbf{x})] - \mathbb{E}_{\mathbf{x} \in D_\mathrm{out}}[-\ln p(\mathbf{x})] \\
&= \mathbb{E}_{\mathbf{x} \in D_\mathrm{in}}[- \ln \hat{p}(\mathbf{x})]  - \mathbb{E}_{\mathbf{x} \in D_\mathrm{out}}[-\ln \hat{p}(\mathbf{x})]  
\label{eq:x_d_0}
\end{align}
Note that the normalization constant $Z$ cancels out due to training with outliers, as detailed in the Appendix.
In practice, we use a simplified loss that is an upper bound of the above expression ($L_{\mathbf{x}}^{\mathrm{UB}} \geq L_{\mathbf{x}}$):
\begin{equation}
    L_{\mathbf{x}}^{\mathrm{UB}}(\theta) = \, \mathbb{E}_{\mathbf{x}, y \in D_\mathrm{in}}[-\mathbf{s}_y]  + \, \mathbb{E}_{\mathbf{x} \in D_\mathrm{out}}[\underset{y'}{\text{LSE}}(\mathbf{s}_{y'})].
\label{eq:x_d}
\end{equation}
We observe that $L_\text{cls}$ and $L_\textbf{x}^{\mathrm{UB}}$ have a shared loss term.
Recall that training data likelihood only on inliers \cite{du19neurips,grathwohl20iclr} would require MCMC sampling, which is infeasible in our context. 
Unnormalized likelihood could also be trained through score matching \cite{song19neurips}.
However, this would preclude hybrid modelling due to having to train on noisy inputs.
Consequently, it appears that  the proposed training approach is a method of choice in our context.
Comparison of the discriminative loss (\ref{eq:cls}) and the generative upper bound (\ref{eq:x_d}) reveals that the standard classification loss is well aligned with the upper bound in inlier pixels.
The proof of inequality $L_{\mathbf{x}}^{\mathrm{UB}} \geq L_{\mathbf{x}}$ is in the Appendix.
\begin{figure*}
    \centering
    \includegraphics[width=\linewidth]{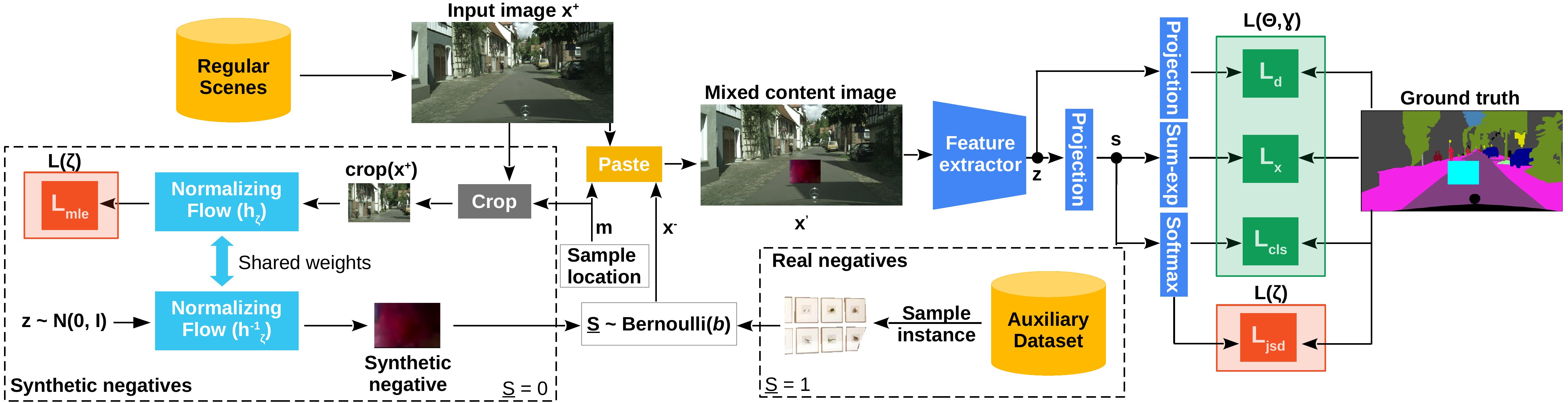}
    \caption{
    Fine-tuning procedure for the proposed DenseHybrid model.
    We construct mixed-content images by pasting negatives into inlier images according to (\ref{eq:pasting}).
    The negative training data can be sourced from an auxiliary real dataset (Sec.\ \ref{sec:oss_train_aux}), from a jointly trained normalizing flow (Sec.\ \ref{sec:oss_train_syn}), or from both sources according to $b$ from (\ref{eq:neg_mix}).
    Mixed-content images are fed to the open-set model that produces three dense outputs: the closed-set class posterior, unnormalized data likelihood, 
and dataset posterior. 
The model is optimized according to the compound loss (\ref{eq:final_loss}).
    In the case of synthetic negatives ($S=0$), the normalizing flow optimizes the loss (\ref{eq:flow_total}).
    }
    \label{fig:ratio_train}
\end{figure*}

\noindent
\textbf{Dataset posterior.}
The dataset-posterior head  $P(d_\mathrm{in}|\mathbf{x})$ 
 requires a discriminative loss that distinguishes the inliers $\mathbf{x} \in D_\mathrm{in}$ from the outliers $ \mathbf{x} \in D_\mathrm{out}$ 
\cite{bevandic22ivc}:
\begin{align}
\label{eq:d_x}
    L_{\mathbf{d}}(\theta, \gamma) = &-\mathbb{E}_{\mathbf{x} \in D_\mathrm{in}}[ \ln P(d_\mathrm{in}|\mathbf{x})] \nonumber \\ &- \mathbb{E}_{\mathbf{x} \in D_\mathrm{out}}[ \ln ( 1 - P(d_\mathrm{in}|\mathbf{x}))].
\end{align}
\noindent
\textbf{Compound loss.} Our final compound loss aggregates $L_{\mathrm{cls}}$, $L_{\mathbf{x}}^{\mathrm{UB}}$ and $L_{\mathbf{d}}$:
\begin{align}
\label{eq:final_loss}
    L(\theta, \gamma) = &- \mathbb{E}_{\mathbf{x},y \in D_\mathrm{in}}[\ln P(y|\mathbf{x}) + \ln P(d_\mathrm{in}|\mathbf{x})] \nonumber \\
    &- \mathbb{E}_{\mathbf{x} \in D_\mathrm{out}}[ \ln(1 -  P(d_\mathrm{in}|\mathbf{x})) - \ln \hat{p}(\mathbf{x})].
\end{align}
In practice, we use a modulation hyperparameter for every loss component.
More on the hyperparameters, together with the complete derivation, can be found in the Appendix.

Figure \ref{fig:ratio_train} illustrates the training of our open-set segmentation models.
The figure shows that we prepare mixed-content training images $\mathbf{x}'$ by pasting negative patches $\mathbf{x}^- \in D_\mathrm{out}$ into regular training images $\mathbf{x}^+ \in D_\mathrm{in}$:
\begin{equation}
    \mathbf{x}' = (\mathbf{1}-\mathbf{m}) \cdot \mathbf{x}^+ +  \mathrm{pad}(\mathbf{x}^-,\mathbf{m}).
    \label{eq:pasting}
\end{equation}
The binary mask $\mathbf{m}$ identifies negative pixels within the mixed-content image $\mathbf{x}'$.
Semantic labels of negative pixels are set to void. 
The resulting mixed-content image $\mathbf{x}'$ is fed to the segmentation model that produces pre-logits $\mathbf{z}$ and logits $\mathbf{s}$. 
We recover the class posterior, unnormalized likelihood, and dataset posterior 
as explained in Sec.\ \ref{sec:prob_view}, and perform the training
with respect to the loss (\ref{eq:final_loss}).

\subsection{Open-set Training with Synthetic Negative Data}
\label{sec:oss_train_syn}

Training anomaly detectors on real negative training data
may result in over-optimistic performance estimates
due to non-empty intersection between 
the training negatives and test anomalies.
This issue can be addressed by replacing real negative training data with samples from a suitable generative model  \cite{lee18iclr,grcic21visapp,kong22tpami,zhao23pami}.
The generative model can be trained to generate synthetic samples that encompass the inlier distribution \cite{lee18iclr}. 
The required learning signal can be derived from discriminative predictions \cite{lee18iclr,grcic21visapp,zhao23pami} 
or provided by an adversarial module \cite{besnier21iccv}. 
Anyway, replacing real negative data with synthetic counterparts requires joint training of the generative model. 
We choose a normalizing flow \cite{grcic21neurips} due to fast training, good distributional coverage, and fast generation at varying spatial dimensions \cite{grcic24sensors}.

We train the normalizing flow $p_\zeta$ according to the data term and boundary-attraction term.
The data term $L_\mathrm{mle}$ corresponds to image-wide negative log-likelihood of random crops from inlier images $\mathbf{x}^+$:
\begin{equation}
    L_{\mathrm{mle}}(\zeta) = - \mathbb{E}_{\mathbf{x}^+ \in D_\mathrm{in}}[\ln p_\zeta(\text{crop}(\mathbf{x}^+, \mathbf{m}))].
\end{equation}
The crop notation mirrors the pad notation from (\ref{eq:pasting}).
Random crops vary in spatial resolution.
This term aligns the generative distribution with the distribution of the training data.
It encourages coverage of the inlier distribution 
assuming sufficient capacity of the generative model.

The boundary-attraction term $L_\mathrm{jsd}$ \cite{grcic24sensors} corresponds to negative Jensen-Shannon divergence between the class-posterior and the uniform distribution across all generated pixels.
This term pushes the generative distribution towards the periphery of the inlier distribution where the class posterior should have a high entropy.
Note that gradients of this term must propagate through the entire segmentation model in order to reach the normalizing flow.
Hence, the flow is penalized when the generated sample yields high softmax confidence. 
This signal pushes the generative distribution away from high-density regions of the input space \cite{lee18iclr}.
The total normalizing flow loss modulates the contribution of the boundary term with hyperparameter $\lambda$: 
\begin{equation}
\label{eq:flow_total}
    L(\zeta) = L_{\mathrm{mle}}(\zeta) + \lambda \cdot L_{\mathrm{jsd}}(\zeta; \theta)
\end{equation}
Optimization of (\ref{eq:flow_total}) enforces the generative distribution to encompass the inlier distribution. 
Note that our normalizing flow can never match the diversity of images from a real dataset such as ADE20k.
It would be unreasonable to expect a generation of a sofa after training on traffic scenes. 
Still, if the flow succeeds to learn the boundary of the inlier distribution, then DenseHybrid will be inclined to associate
all off-distribution datapoints 
with  low $s_H$.

Details of the training procedure are illustrated in Figure \ref{fig:ratio_train}.
We sample the normalizing flow by \textit{i)} selecting a random spatial resolution (H,W) from a predefined interval,
\textit{ii)} sampling a random latent representation
$\mathbf{z} \sim \mathcal{N}(0,\mathrm{I}_{HW})$, and \textit{iii)} feeding $\mathbf{z}$ to the flow so that $\mathbf{x}^- = h_\zeta^{-1}(\mathbf{z})$.
We again craft a mixed-content image $\mathbf{x}'$ by pasting the synthesized negative patch $\mathbf{x}^- \sim p_\zeta$ into the regular training image $\mathbf{x}^+ \in D_{\text{in}}$ according to (\ref{eq:pasting}), perform the forward pass, determine $L_\mathrm{cls}$, $L_\mathbf{d}$, $L_\mathbf{x}$, and $L_\mathrm{jsd}$, and recover the training gradients by backpropagation.
We now take the deleted inlier patch $\mathbf{x}^{+}_\mathbf{s}$, perform inference with the normalizing flow
($\mathbf{z} = h_\zeta(\mathbf{x}^{+}_\mathbf{s}$)) and accumulate gradients of $L_\mathrm{mle}$ before performing a model-wide parameter update.

We can also source the negative content from a mixture of real and synthetic samples.
Then, the amount of data from each source is modulated by hyperparameter $b \in [0, 1]$. The probability of sampling a real negative equals $b$, while the probability of sampling a synthetic negative equals $1-b$.
Hence, the distribution of mixed negatives $p_\text{neg}$ is:
\begin{equation}
\label{eq:neg_mix}
    p_\text{neg}(\mathbf{x}^-) = b \cdot p_\text{out}(\mathbf{x}^-) +  (1 - b) \cdot p_\zeta(\mathbf{x}^-), \; b \in [0, 1] 
\end{equation}
Sampling $p_\text{neg}$ proceeds by first choosing the source, which corresponds to sampling a Bernoulli distribution $\mathcal{B}(b)$.
Then, the negative is generated by sampling the selected source.

\section{Experimental setup}

Evaluation of dense anomaly detection and open-set segmentation
requires specialized datasets and benchmarks (Sec.\ \ref{sec:bandd}).
None of the existing metrics can quantify the gap 
between open-set and closed-set segmentation performance.
We address this need by proposing 
a novel Open-IoU metric (Sec.\ \ref{sec:openiou}).
The remaining implementation details 
are described in the Appendix. 
Our code will be publicly available within the DenseHybrid repository \cite{grcic22code} upon acceptance.

\subsection{Benchmarks and Datasets}
\label{sec:bandd}
We evaluate performance on benchmarks for dense anomaly detection and open-set segmentation.
The Fishyscapes benchmark \cite{blum21ijcv} consists of a subset of the LostAndFound dataset \cite{pinggera16iros} (FS LAF) and  Cityscapes validation images with pasted anomalies (FS Static).
The SegmentMeIfYouCan dataset (SMIYC) \cite{chan21neuripsd} collects carefully selected images from the real world and groups them into the AnomalyTrack (large anomalies) and ObstacleTrack (small anomalies on the road surface).
Moreover, the benchmark includes a selection of  images from LostAndFound \cite{pinggera16iros} where the lost objects do not correspond to the Cityscapes taxonomy (LAF-noKnown).
We report only anomaly detection 
on these benchmarks 
since they supply only binary labels 
and thus preclude open-set evaluation. 
We validate performance on Cityscapes while reinterpreting a subset of ignore classes as the unknown class \cite{kong22tpami}.
The StreetHazards dataset \cite{hendrycks22icml} is a synthetic dataset that enables smooth anomaly injection and low-cost label extraction.
Consequently, the dataset contains K+1 labels, making it suitable for dense open-set evaluation.
We also validate open-set segmentation on crowdsourced photos from COCO val \cite{lin14eccv}.
We set the 20 Pascal VOC classes as inliers while the remaining classes are considered unknown.
In the first case, we train on images from augmented VOC 2012 \cite{hariharan11iccv} and consider the remaining 113 classes as unknown unknowns.
We set all background training pixels to the mean pixel to prevent leakage of anomalous semantic content to the inlier representations.
We call this setup Pascal-COCO.
In the second case, we set the remaining 60 COCO thing classes as unknowns and
conduct training on images from COCO train split that contain at least one known class and no unknown classes.
Furthermore, we ignore all pixels of the stuff classes.
We call this setup COCO20/80.
In both cases, we observe a covariate shift between the training and test datasets which makes these setups quite challenging.

\subsection{Measuring Open-set Performance}
\label{sec:openiou}
Previous work evaluates open-set segmentation through anomaly detection \cite{pinggera16iros,chan21neuripsd} and closed-set segmentation \cite{blum21ijcv}.
The observed drop in closed-set performance is usually negligible and is explained by the allocation of model capacity for anomaly detection.
However, we will show that the impact of anomalies onto segmentation performance can be clearly characterized only in the open-set setup.
More precisely, we shall take into account false positive  semantic predictions at anomalies as well as false negative semantic predictions due to false anomaly detections.

We propose a novel evaluation procedure for open-set segmentation.
Our procedure starts by thresholding the anomaly score
so that it yields 95\% TPR anomaly detection on held-out data. 
Then, we override the classification in pixels which score higher than the threshold.
This yields a recognition map with $K+1$ labels.
We assess open-set segmentation performance
according to a novel metric that we term open-mIoU.
We compute open-IoU for the $k$-th class as follows:
\begin{equation}
    \text{open-IoU}_k = \frac{\mathrm{TP}_k}{\mathrm{TP}_k + \mathrm{FP}^\text{os}_k + \mathrm{FN}^\text{os}_k}, \quad\mathrm{where}
\end{equation}
\begin{equation}
\mathrm{FP}^\text{os}_{k} = \sum_{i=1, i\neq k}^{K+1} \mathrm{FP}_k^i, \quad \mathrm{FN}^\text{os}_{k} = \sum_{ i=1, i\neq k}^{K+1} \mathrm{FN}_k^i .
\end{equation}
Different than the standard IoU formulation, open-IoU takes into account false predictions due to imperfect anomaly detection.
In particular, a prediction of class $k$ at an outlier pixel (false negative anomaly detection) counts as a false positive for class $k$.
Furthermore, a prediction of class K+1 at a pixel labelled as inlier class  $k$ (false positive anomaly detection) counts as a false negative for class $k$.
Note that we still average open-IoU over $K$ inlier classes.
Thus, a recognition model with perfect anomaly detection gets assigned the same performance as in the closed world.
This property would not be preserved if we averaged open-IoU over K+1 classes.
Hence, a comparison between mIoU and open-mIoU quantifies the gap between the closed-set and open-set performance, unlike the related metrics \cite{sokolova09ipm,kong22tpami}.

\begin{table*}[ht]
\centering
\caption{Dense anomaly detection on SegmentMeIfYouCan \cite{chan21neuripsd} and Fishyscapes \cite{blum19iccvw}. Aux data denotes training on real negatives, 
while Img rsyn.\ denotes image resynthesis.}
\label{tbl:smiyc}
\begin{footnotesize}
\begin{tabular}{lccccccccccccc}
\multirow{3}{*}{Method} & & & \multicolumn{6}{c}{SegmentMeIfYouCan \cite{chan21neuripsd}} &  \multicolumn{5}{c}{Fishyscapes \cite{blum21ijcv}} \\
 & \multicolumn{1}{c}{\multirow{2}{*}{Aux}} & \multicolumn{1}{c}{\multirow{2}{*}{Img}} & \multicolumn{2}{c}{AnomalyTrack} & \multicolumn{2}{c}{ObstacleTrack}
& \multicolumn{2}{c}{LAF-noKnown} & \multicolumn{2}{c}{FS LAF} & \multicolumn{2}{c}{FS Static} & CS val\\
 & \multicolumn{1}{c}{data} & \multicolumn{1}{c}{rsyn.} & \multicolumn{1}{c}{AP} & \multicolumn{1}{c}{$\mathrm{FPR}_{95}$} & AP & \multicolumn{1}{c}{$\mathrm{FPR}_{95}$} & AP & \multicolumn{1}{c}{$\mathrm{FPR}_{95}$}  & AP & \multicolumn{1}{c}{$\mathrm{FPR}_{95}$} & AP & \multicolumn{1}{c}{$\mathrm{FPR}_{95}$} & $\overline{\mathrm{IoU}}$ \\ \hline 

Image Resyn. \cite{lis19iccv} & \xmark& \cmark& \textbf{52.3} & \textbf{25.9} & 37.7 & 4.7 & 57.1 & 8.8 & 5.7 & 48.1 & 29.6 & 27.1 & 81.4  \\
Road Inpaint. \cite{lis20arxiv} & \xmark& \cmark & - & -  & 54.1 & 47.1 & 82.9 & 35.8  & - & - & - & - & -  \\
Max softmax \cite{hendrycks17iclr} & \xmark&  \xmark & 28.0 & 72.1 & 15.7 & 16.6  &  30.1 & 33.2 &  1.8 & 44.9 & 12.9 & 39.8 & 80.3 \\
MC Dropout \cite{kendall17nips} & \xmark& \xmark & 28.9 & 69.5 & 4.9 & 50.3 & 36.8 & 35.6 & - & - & - & - & - \\
ODIN \cite{liang18iclr} & \xmark&  \xmark & 33.1 & 71.7 & 22.1 & 15.3 & 52.9 & 30.0 & - & - & - & - & -  \\
SML \cite{jung21iccv} & \xmark & \xmark & - & - & - & - & - & -    & 31.7 & 21.9 & 52.1 & 20.5 & -\\
Embed.\ Dens.\ \cite{blum21ijcv} & \xmark&  \xmark & 37.5 & 70.8 & 0.8 & 46.4 & 61.7 & 10.4 & 4.3 & 47.2 & \textbf{62.1} & 17.4 & 80.3\\
JSRNet \cite{vojir21iccv} & \xmark&  \xmark & 33.6& 43.9 & 28.1 & 28.9 & 74.2 & 6.6 & - & - & - & - & - \\
SynDenseHybrid (ours) & \xmark&  \xmark & \textbf{51.5} & \textbf{33.2}  & \textbf{64.0} & \textbf{0.6} & \textbf{78.8} & \textbf{1.1} & \textbf{51.8} & \textbf{11.5} & 54.7 & \textbf{15.5} & 79.9 \\[.5em]
SynBoost \cite{biase21cvpr} & \cmark& \cmark & 56.4 & 61.9 & 71.3 & 3.2 & 81.7 & 4.6 & 43.2 & 15.8 & 72.6 & 18.8 & 81.4 \\
Prior Entropy \cite{malinin18nips} & \cmark & \xmark & - & - & - & - & - & - & 34.3 & 47.4 & 31.3 & 84.6 & 70.5 \\
OOD Head \cite{bevandic19gcpr} & \cmark & \xmark & - & - & - & - & - & -  & 31.3 & 19.0 & \textbf{96.8} & \textbf{0.3} & 79.6 \\
Void Classifier \cite{blum21ijcv} & \cmark& \xmark & 36.6 & 63.5 & 10.4 & 41.5 & 4.8 & 47.0 & 10.3 & 22.1 & 45.0 & 19.4 & 70.4 \\
Dirichlet prior \cite{malinin18nips} & \cmark& \xmark & - & -& - & - & - & - &  34.3 & 47.4 & 84.6 & 30.0 & 70.5\\ 
DenseHybrid (ours) & \cmark&  \xmark & \textbf{78.0} & \textbf{9.8}  & \textbf{87.1} & \textbf{0.2} & 78.7 & \textbf{2.1} & \textbf{43.9} & \textbf{6.2} & 72.3 & \textbf{5.5} & 81.0 \\
\end{tabular}
\end{footnotesize}
\end{table*}

\begin{figure}[ht]
    \centering
    \includegraphics[width=\linewidth]{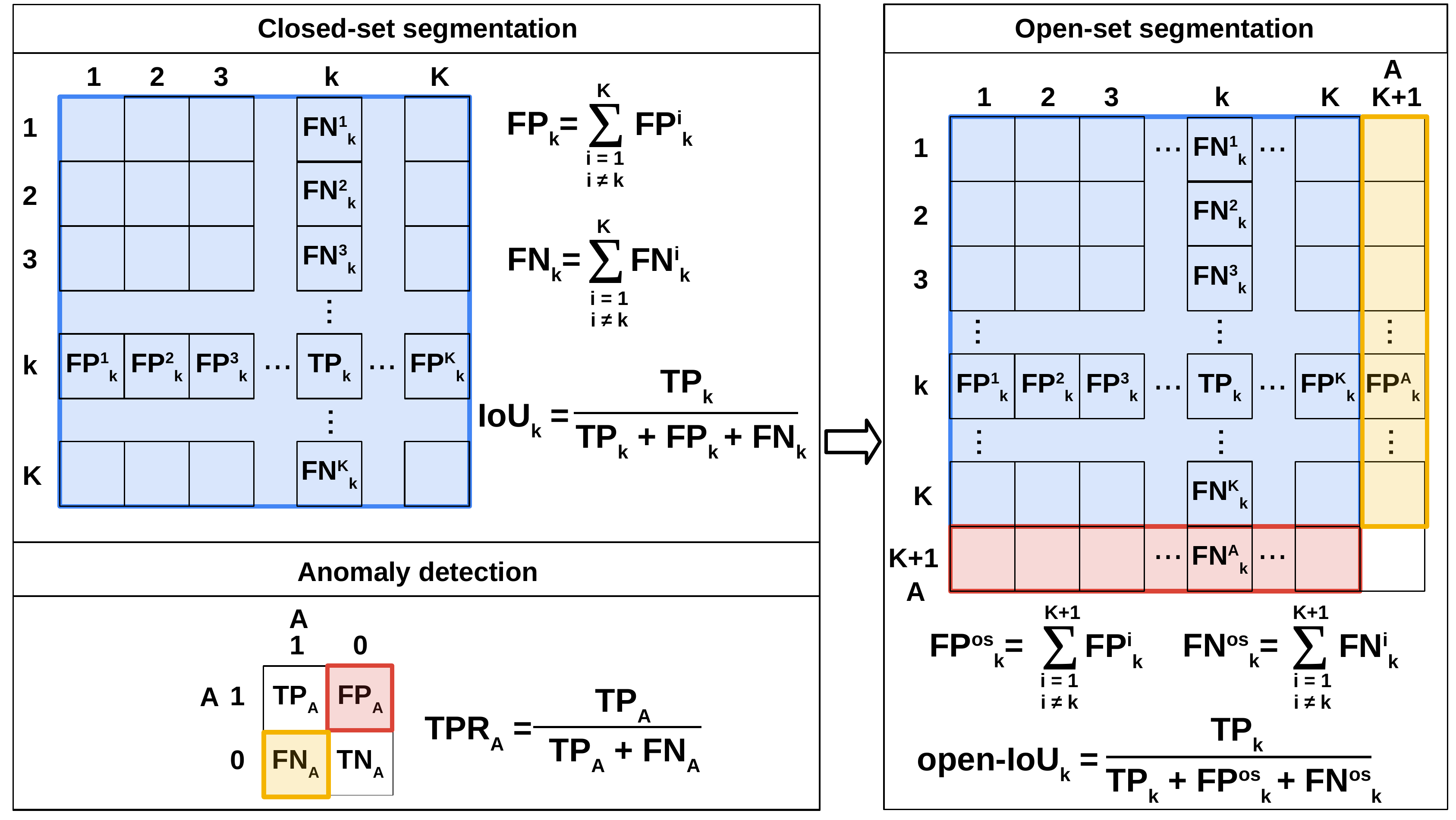}
    \caption{
    We extend the standard closed-set metric (top-left)
with a novel open-set metric (right).
Open-IoU takes into account 
false positive  semantics at undetected anomalies
as well as false negative semantics 
due to false positive anomalies.
The proposed open-mIoU metric quantifies dense
recognition performance in the presence of anomalies.}
    \label{fig:conf_mat}
\end{figure}

Figure \ref{fig:conf_mat} compares the considered closed-set (top left, $\text{IoU}_k$) and open-set (right, $\text{open-IoU}_k$) metrics.
Imperfect anomaly detection impacts recognition performance through increased false positive and false negative semantics (designated in yellow and red, respectively).
Difference between closed-set mIoU and open-mIoU reveals the performance gap due to inaccurate anomaly detection.

\section{Experimental results}

We evaluate DenseHybrid in dense anomaly detection (Sec.\ \ref{sec:res_anomdet}) and open-set segmentation (Sec.\ \ref{sec:res_rec})
after training with and without real negative data.
Further experiments present ablations (Sec.\ \ref{sec:ablations})
and evaluate mixed negatives (Sec.\ \ref{sec:syn_neg_source}).


\subsection{Dense Anomaly Detection in Open-set Setups}
\label{sec:res_anomdet}

Table \ref{tbl:smiyc} presents dense anomaly detection performance on SMIYC \cite{chan21neuripsd} and Fishyscapes \cite{blum19iccvw}.
We include our models trained on real negative data (DenseHybrid) and on synthetic negatives (SynDenseHybrid).
Following \cite{chan21iccv,vojir21iccv,chan21neuripsd}, we use
the standard Cityscapes-trained DeepLabV3+ \cite{zhu19cvpr} for Fishyscapes.
For SMIYC, we train LDN-121 \cite{kreso21tits} on images from Cityscapes, Vistas, and Wilddash.
DenseHybrid outperforms contemporary approaches 
on SMIYC Anomaly, SMIYC Obstacle, 
and FS LAF 
by a wide margin.
Furthermore, it achieves the best $\mathrm{FPR}_{95}$ on 
SMIYC LAF-noKnown and FS Static.

Among methods that do not train on real negatives,
SynDenseHybrid prevails on
on SMIYC Obstacle, SMIYC LAF-noKnown,
and FS LAF. 
Furthermore, it achieves the best FPR on FS Static
and ranks second on SMIYC anomaly and FS Static AP.
All our performance estimates use standard performance metrics of the particular datasets.
Our performance metrics on FS LAF would increase if we considered only the road pixels as in \cite{vojir21iccv}.
The rightmost column of the table indicates that our fine-tuning protocol exerts a negligible impact on closed-set performance.
However, the next section will show that the impact of anomaly detection on final recognition performance is more significant than what can be measured with closed-set metrics.

We also validate our 
 method by considering a subset of Cityscapes void classes as the unknown class.
More precisely, we consider all void classes except 'unlabeled', 'ego vehicle', 'rectification border', 'out of roi' and 'license plate' as unknowns during validation \cite{kong22tpami}.
Table \ref{table:opengan_cs} compares performance according to the AUROC (AUC) metric.
SynDenseHybrid outperforms all previous works.
Most notably, it outperforms the previous SotA \cite{kong22tpami} by four percentage points.
To offer fair comparison with previous work, we do not report results when training on real negative data.
\begin{table}[ht]
\centering
\caption{Anomaly detection on Cityscapes val with a subset of ignore classes considered as unknowns \cite{kong22tpami}.
}
\label{table:opengan_cs}
\begin{tabular}{lclc}
Method & AUC & Method & AUC \\ \hline
MSP \cite{hendrycks17iclr} & 72.1 & GDM \cite{lee18nips} & 74.3 \\
Entropy \cite{steinhardt16neurips} & 69.7 & GMM & 76.5\\
OpenMax \cite{bendale16cvpr} & 75.1 &  K+1 classifier & 75.5 \\
C2AE \cite{oza19cvpr} & 72.7 &  OpenGAN-O \cite{kong22tpami} & 70.9 \\
ODIN \cite{liang18iclr} & 75.5 &  OpenGAN \cite{kong22tpami} & 88.5 \\
MC dropout \cite{kendall17nips} & 76.7  & SynDenseHybrid (ours) &   \textbf{92.9}\\
\end{tabular}
\end{table}

\begin{figure}[ht]
    \centering
    \includegraphics[width=\linewidth]{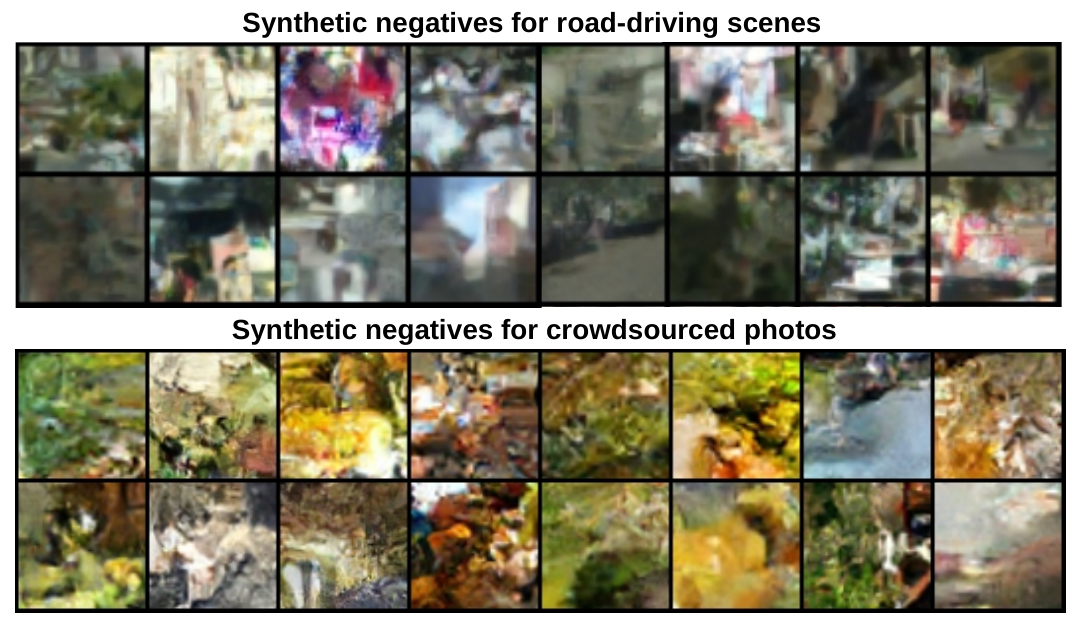}
    \caption{Dataset-specific synthetic negatives sampled from our normalizing flows (cf.\ Sec.\ \ref{sec:oss_train_syn}). During the training, we sample the normalizing flow at different resolutions to mimic anomalies of different sizes.
}
    \label{fig:negatives}
\end{figure}

Figure \ref{fig:negatives} shows synthetic negatives produced by the training setup from Sec.\ \ref{sec:oss_train_syn}.
Samples vary in spatial resolution and lack meaningful visual concepts.
Yet, training our open-set model on such samples yields only slightly worse performance
than when training on real negative data.

\subsection{Open-set Segmentation
}
\label{sec:res_rec}

We consider open-set performance 
according to mean $F_1$ ($\overline{F_1}$) score and the proposed open-mIoU (o$\overline{\mathrm{IoU}}$) metric.
Table \ref{table:osr_sh} presents performance evaluation on the StreetHazards dataset\footnote{Note that Table \ref{table:osr_sh} does not list ObsNet \cite{besnier21iccv} since they aim to detect classification errors instead of anomalies.}.
The left part of the table considers anomaly detection while the right part considers closed-set and open-set segmentation.
Our method outperforms contemporary approaches in anomaly detection both with and without training on real negative data.
Furthermore, our method achieves the best open-set performance (columns o$\overline{\mathrm{IoU}}$ and $\overline{F_1}$) despite moderate capacity of LDN-121 ($\overline{\mathrm{IoU}}$ column).
The last column quantifies the performance gap 
between closed-set and open-set performance
as the difference between IoU and oIoU.
Our method achieves the least performance gap of around 18 percentage points. 
Nevertheless, an ideal model would deliver equal open-set and closed-set metrics.
Hence, we conclude that the current state of the art is incapable to deliver closed-set performance in open-set setups.
Researchers should continue the effort to diminish this gap in order to improve the safety of recognition systems in the real world.

\begin{table}[ht]
\begin{center}
\caption{Performance evaluation on StreetHazards \cite{hendrycks22icml}.
We evaluate anomaly detection (Anomaly), closed-set 
(Clo.) and open-set segmentation (Open-set), as well as the open-set performance gap (Gap).
}
\label{table:osr_sh}
\setlength{\tabcolsep}{5pt}
\begin{tabular}{lccccccc}
\multirow{2}{*}{Method} &  \multicolumn{3}{c}{Anomaly} & \multicolumn{1}{c}{Clo.} & \multicolumn{2}{c}{Open-set} & \multirow{2}{*}{Gap}\\
    &  AP         & $\text{FPR}_{95}$ & \multicolumn{1}{c}{AUC}    & \multicolumn{1}{c}{$\overline{\mathrm{IoU}}$} & $\overline{F_1}$ & $\overline{\text{oIoU}}$  &  \\ \hline 
SynthCP \cite{xia20eccv} &   9.3           & 28.4  & 88.5      & - & -& - & - \\
Dropout \cite{kendall17nips} &    7.5           & 79.4  & 69.9   & - & - & - & - \\
TRADI \cite{franchi20eccv} &  7.2           & 25.3 & 89.2    & - & - & - & -  \\
SO+H \cite{grcic21visapp}&  12.7  & 25.2  & 91.7 & 59.7 & - & - & - \\
DML \cite{cen21iccv}  & 14.7  & 17.3  &  93.7  &  -  & - & - & -\\
MSP \cite{hendrycks17iclr} &   7.5   &   27.9 &   90.1 &  65.0  & 46.4  & 35.1  & 29.9\\
ODIN \cite{liang18iclr}&     7.0       & 28.7  &   90.0    & 65.0 & 41.6 & 28.8  & 36.2\\
ReAct \cite{sun21neurips} &  10.9  & 21.2 & 92.3 &  62.7  & 46.4 & 34.0  &  28.7\\
SynDnsHyb &  \textbf{19.7}  &  \textbf{17.4}  &  \textbf{93.9} &  61.3 & \textbf{50.6} & \textbf{37.3} & \textbf{24.0}\\[.5em]

Energy \cite{liu20neurips}&   12.9  & 18.2 & 93.0  & 63.3  & 50.4 &  42.7 & 29.9 \\
OE \cite{hendrycks19iclr} &  14.6   &  17.7 & 94.0  & 61.7 & 56.1 & 43.8  & 17.9 \\
OH \cite{bevandic19gcpr}  &  19.7 &  56.2 & 88.8   &  \textbf{66.6} & - & 33.9  & 32.7 \\
OH*MSP \cite{bevandic22ivc} &  18.8  & 30.9 & 89.7  &   \textbf{66.6} & -  & 43.6 & 23.0 \\
DenseHybrid &  \textbf{30.2}  & \textbf{13.0} & \textbf{95.6}  & 63.0 & \textbf{59.7} & \textbf{45.8} & \textbf{17.2} \\
\end{tabular}
\end{center}
\end{table}

\begin{figure}[ht]
    \centering
    \includegraphics[width=\linewidth]{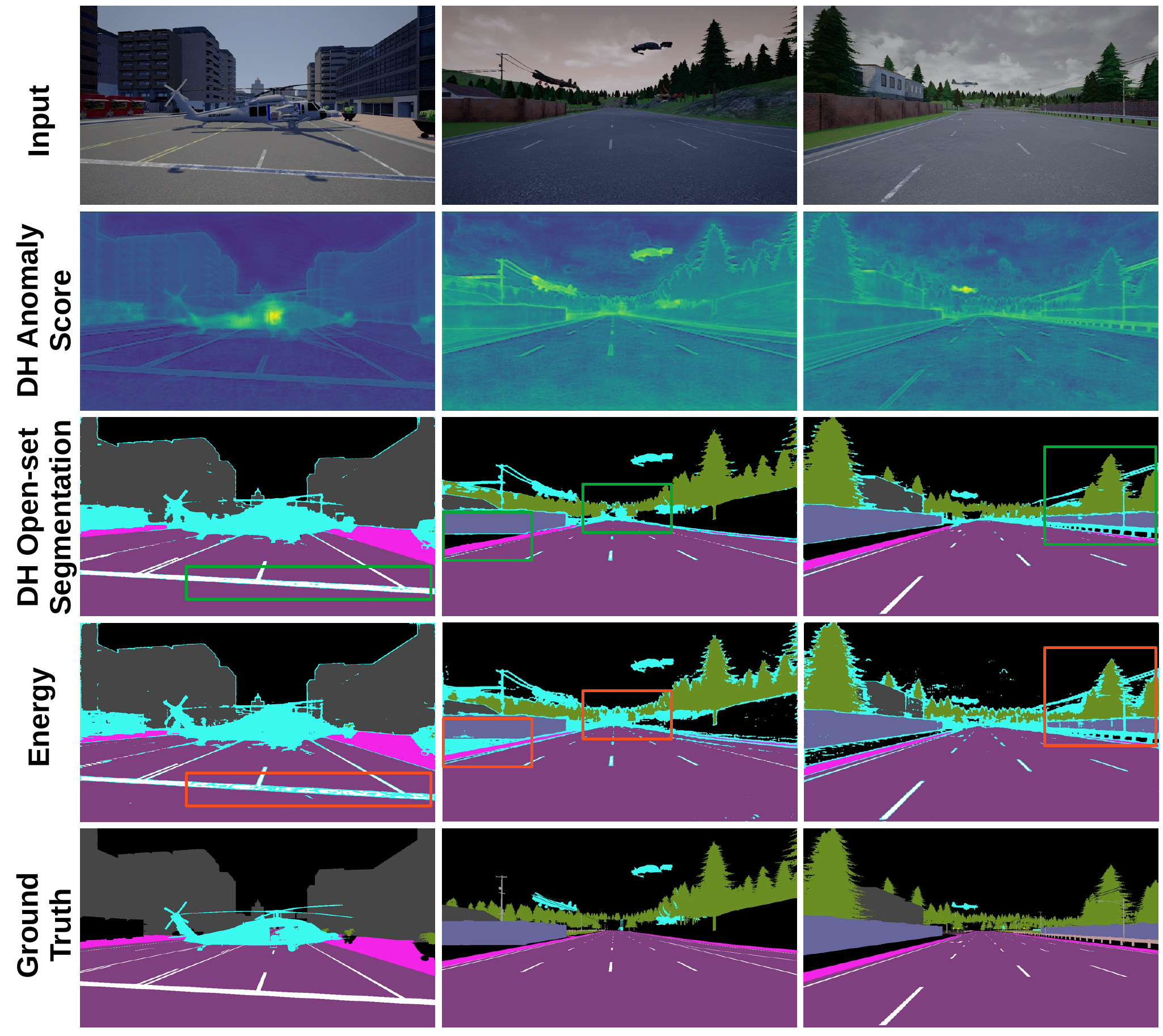}
    \caption{Qualitative open-set segmentation performance on StreetHazards.
    DenseHybrid (rows 2 and 3) delivers a more accurate open-set performance with respect to the energy-based approach \cite{liu20neurips} (row 4), as denoted with red rectangles. Please zoom in for a better view.}
    \label{fig:sh_results}
\end{figure}

Figure \ref{fig:sh_results} visualizes qualitative open-set segmentation performance on StreetHazards test.
Our hybrid model accurately combines dense anomaly detection (second row) with closed-set segmentation, and delivers open-set segmentation (third row).
Contemporary energy-based approach \cite{liu20neurips} yields more false positives at TPR=95\% (fourth row).


Table\ \ref{table:oss_general_scenes} presents open-set segmentation performance on crowdsourced photos from the COCO dataset.
We use Segmenter \cite{strudel21iccv} with ViT-B/16 as the closed-set classifier.
Our hybrid open-set method outperforms previous approaches when training with real negatives from ADE20K \cite{hendrycks19iclr,liu20neurips}
as well as with synthetic negatives
\cite{hendrycks22icml,vaze22iclr,liang18iclr}.
Closed-set models reach more than 90\% mIoU in the case of Pascal-COCO and over 75\% in the case of COCO20/80, but open-IoU peaks at 41\% and 16.3\%.
Analysis of false positives reveals that the task is hard due to large intra-class variation (e.g.\ different species of potted plants).
Moreover, some unknown classes have similar appearance to known classes (eg.\ unknown \textit{zebra} and known \textit{horse}).
Finally, the benchmark has high openness \cite{scheirer12tpami}: there are more than $3\times$ unknown than known classes.
Interestingly, synthetic negatives prevail in the case of COCO20/80.
This indicates that ADE20k negatives may not be an adequate negative dataset for this setup.
We show qualitative examples of our open-set segmentation performance in the Appendix.
These examples show that many mistakes of our model coincide with labelling errors.
\begin{table}[ht]
 \setlength{\tabcolsep}{3pt}
\begin{center}
\caption{
Open-set segmentation on crowdsourced photos from COCO val.}
\label{table:oss_general_scenes}
\begin{tabular}{lcccccccc}
\multirow{2}{*}{Method} & \multicolumn{4}{c}{Pascal - COCO}  & \multicolumn{4}{c}{COCO 20/80} \\
  & $\text{FPR}_{95}$ & $\overline{F_1}$ & $\overline{\text{IoU}}_{21}$ & $\overline{\text{oIoU}}$ & $\text{FPR}_{95}$ & $\overline{F_1}$ & $\overline{\text{IoU}}_{21}$ & $\overline{\text{oIoU}}$ \\ \hline
ML \cite{hendrycks22icml,vaze22iclr}  & 56.5  & 22.0 & 16.5 & 13.2 & 63.5 & 15.6 & 11.6 & 9.8 \\
ODIN \cite{liang18iclr}  & 66.1 & 13.3 & 10.0 & 6.5 & 61.4 & 15.8 & 12.2  & 10.5\\
SynDnsHyb &  \textbf{46.8}  & \textbf{48.4} & \textbf{35.3} & \textbf{33.9} & \textbf{59.7} & \textbf{24.2} & \textbf{17.8} & \textbf{16.3}\\[.5em]

Energy \cite{liu20neurips} &  54.6  & 19.3 & 14.8 & 11.4 & 65.7 & 14.7 & 10.7 & 8.9\\
OE \cite{hendrycks19iclr} &  46.0 & 49.1  & 36.5 & 34.1 & 68.7 & 13.5 & 9.5 & 7.7 \\
DenseHybrid &  \textbf{43.0} & \textbf{55.3}  & \textbf{42.0} & \textbf{41.0}  & \textbf{61.7} & \textbf{22.8} &  \textbf{16.3} & \textbf{14.7}\\
\end{tabular}
\end{center}
\end{table}

\subsection{Ablating Components of DenseHybrid}
\label{sec:ablations}

Table \ref{table:abl_sx} validates components of our hybrid approach on Fishyscapes val.
The top two sections validate the two DenseHybrid components, $\hat{p}(\mathbf{x})$ and $P(d_\mathrm{in}|\mathbf{x})$, when training on real and synthetic negative data, respectively.
We observe that the hybrid score outperforms unnormalized density which outperforms dataset posterior.
We observe the same qualitative behaviour 
when training on real and synthetic negative data.
The bottom section replaces our unnormalized likelihood with pre-logit likelihood estimates
by a normalizing flow.
The flow is applied point-wise in order to obtain dense likelihood \cite{blum21ijcv}.
This can also be viewed as a generalization of a previous image-wide open-set approach \cite{zhang20eccv} to dense prediction.
We still train on negative data in an end-to-end fashion in order to make the two generative components comparable.
The resulting model delivers good performance on FS Static and poor performance on FS LostAndFound.
We attribute better performance of our unnormalized density (\ref{eq:p_x})
with respect to the point-wise flow
due to 4$\times$ subsampling of the pre-logits to which the flow was fitted. 
Moreover, our unnormalized density 
ensures much faster inference.
\begin{table}[ht]
\begin{center}
\caption{Validation of dense hybrid anomaly detection on Fishyscapes val. Our method outperforms its generative and discriminative components.
}
\label{table:abl_sx}
\setlength{\tabcolsep}{5pt}
\begin{tabular}{lccccc}
\multirow{2}{*}{Anomaly detector} & Neg. & \multicolumn{2}{c}{FS LAF} & \multicolumn{2}{c}{FS Static} \\
  & data & AP & $\text{FPR}_{95}$  & AP & $\text{FPR}_{95}$ \\  \hline
Disc. $(1-P(d_\mathrm{in}|\mathbf{x}))$ &  & 46.5 & 38.3  & 53.5 & 30.9 \\
Gen. $\hat{p}(\mathbf{x})=\text{LSE}(\mathbf{s})$ & Real & 58.2 &  7.3  & 58.0 & 5.3 \\ 
Hyb. $(1-P(d_\mathrm{in}|\mathbf{x}))/\hat{p}(\mathbf{x})$ &  & 60.5 & 6.0  & 63.1 & 4.2 \\[.5em]
Disc. $(1-P(d_\mathrm{in}|\mathbf{x}))$ &  & 30.1 & 35.0  & 48.8 & 39.8 \\
Gen. $\hat{p}(\mathbf{x})=\text{LSE}(\mathbf{s})$ & Syn. & 58.1 &  9.0  & 44.6 & 9.5 \\
Hyb. $(1-P(d_\mathrm{in}|\mathbf{x}))/\hat{p}(\mathbf{x})$ &  & 60.2 & 7.9  & 52.1 & 7.7 \\[.5em]

Gen. flow $p(\mathbf{z})$ & \multirow{2}{*}{Real} & 5.7 &  58.9  & 61.7  & 7.6 \\
Hyb. $(1-P(d_\mathrm{in}|\mathbf{x}))/p(\mathbf{z})$ &  & 6.5 &  46.1  & 65.1  & 6.5 \\
\end{tabular}
\end{center}
\end{table}

\bclr{
Table \ref{tab:oss_ad} shows open-set segmentation performance depending on the choice of the anomaly detector on crowdsourced photos.
Generative and discriminative components of our approach
yield comparable open-set performance,
while their ensemble achieves 
substantial further improvement.
A detailed analysis shows that generative and discriminative detectors are only moderately correlated.
In the case of Pascal-COCO we have $\rho$ = 0.59,  $\alpha$ = $1.22$, $e$ = 1.09, $C_1$ = 0.42, $C_2$ = 0.18, while in the case of COCO20/80 we have $\rho$ = 0.56, $\alpha$ = 1.44, $e$ = 1.22, $C_1$ = 0.7, $C_2$ = 0.04.
Hence, the condition (\ref{eq:hybrid_effectivness_test2}) is satisfied.
Note that it makes no sense to ensemble
two arbitrary anomaly detectors since they are often well-correlated (e.g. max-logit \cite{hendrycks22icml} and free-energy \cite{liu20neurips} have $\rho$=0.98), which again supports our approach.
}
\begin{table}[ht]
\centering
\caption{Validation of DenseHybrid components 
on COCO val.}
\label{tab:oss_ad}
\begin{tabular}{lcccc}
Anomaly & \multicolumn{2}{c}{Pascal-COCO} & \multicolumn{2}{c}{COCO20/80} \\
detector & $\overline{\text{oIoU}}$ & $\overline{F_1}$ & $\overline{\text{oIoU}}$ & $\overline{F_1}$ \\ \hline
Gen. $\hat{p}(\mathbf{x})=\text{LSE}(\mathbf{s})$ & 35.8 & 51.3 & 15.2 & 22.9 \\
Disc. $(1-P(d_\mathrm{in}|\mathbf{x}))$ & 36.3 & 52.8  & 12.0 & 21.8\\
Hyb. $(1-P(d_\mathrm{in}|\mathbf{x}))/\hat{p}(\mathbf{x})$ (ours) & \textbf{41.0} & \textbf{55.3} & \textbf{16.3} & \textbf{24.2}
\end{tabular}
\end{table}

\begin{figure}[ht]
    \centering
    \includegraphics[width=0.98\linewidth]{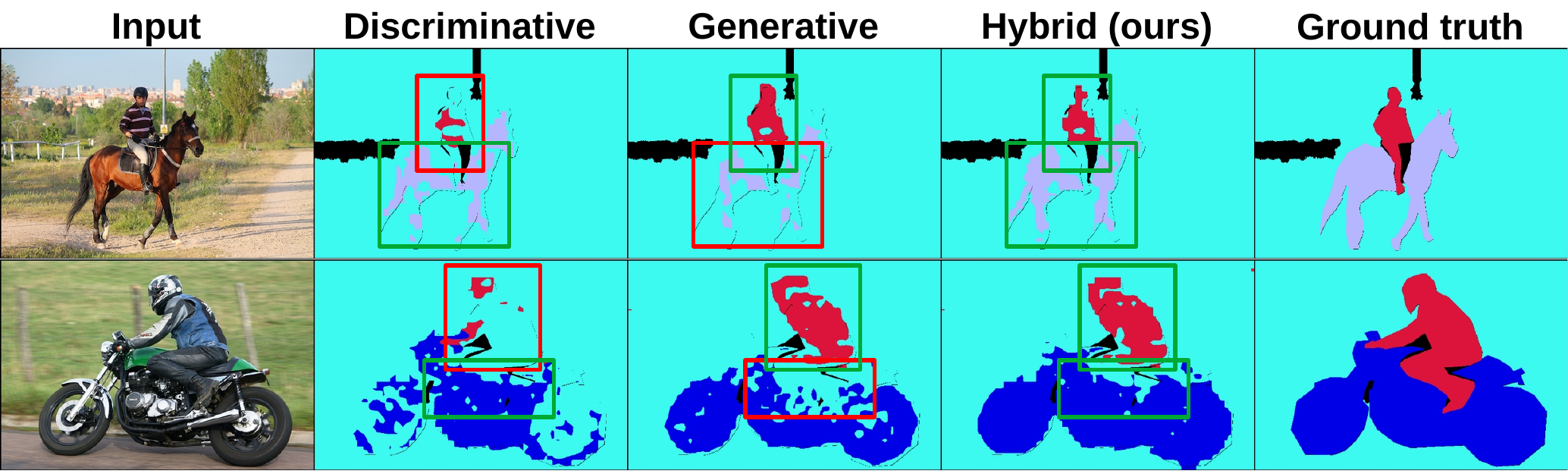}
    \caption{Open-set segmentation on Pascal-COCO with discriminative, generative and hybrid anomaly detection.
    Hybrid anomaly detection yields the lowest FPR95 metric. Red and green boxes indicate abundance and absence of false positive anomalies, respectively.
    }
    \label{fig:abl_hyb_osr}
\end{figure}

Figure \ref{fig:abl_hyb_osr} shows qualitative open-set segmentation experiments, based on different anomaly detectors.
Poorly segmented regions are denoted with red rectangles while green rectangles denote more accurate segmentations.
The Appendix presents a similar qualitative evaluation for road-driving scenes. Please zoom in for a better view.

Table \ref{tbl:syn_source} validates different sources of negative data.
We compare the synthetic negatives from Figure \ref{fig:negatives} with patches of uniform noise, local adversarial attacks \cite{besnier21iccv}, inlier crops \cite{vojir21iccv} as well as 
samples from a jointly trained GAN \cite{lee18iclr}.
We include the average AP over four datasets (last column)
as a metric of overall anomaly detection performance. 
Our synthetic negatives outperform 
all previous approaches in its section
and come close to real negative data.
The bottom section of the table compares real negatives from ADE20k \cite{bevandic22ivc}
with 
10.8k samples from a pre-trained conditional diffusion model \cite{rombach22cvpr}.
We have used prompts of the form "A photograph of \textit{cls}" where \textit{cls} 
stands for a random class description from ADE20k.
Note that diffusion negatives
cannot be directly compared
with other entries due to training
on the huge LAION2B dataset.
\begin{table}[ht]
\setlength{\tabcolsep}{5pt}
\caption{
DenseHybrid performance with different kinds of negative data.
}
\label{tbl:syn_source}
\begin{center}
\begin{tabular}{lccccc}
Source of &  \multicolumn{2}{c}{FS-val} & \multicolumn{2}{c}{SMIYC-val} & \multirow{2}{*}{Average} \\
 negatives & LAF  & Stat. & Anom. & Obs. & \\ \hline
Uniform noise  & 56.9 & 37.4 & 70.5 & 3.5  & 42.1\\
Inlier crops \cite{vojir21iccv}  & 64.3 & 36.2 & 77.2 & 62.8  & 60.1 \\
Loc. Adv. Attacks \cite{besnier21iccv} &   44.5 & 36.8 & 78.9 & 62.2  & 55.6 \\
GAN \cite{lee18iclr} & 60.9 & 38.8 & 72.8 &  44.9 & 54.4 \\
Jointly trained NF & 60.2 & 46.0 &  77.7 &  86.2  & \textbf{67.5} \\[0.5em]

ADE20k-instances \cite{bevandic22ivc} & 63.7 & 68.4 &  76.2 & 86.0  & \textbf{73.6} \\
ADE20k-crops & 62.1 & 47.7 & 76.6 & 74.2 & 65.2 \\
ADE20k-mix & 63.5 & 61.9 & 76.6 & 87.0 & 72.3\\
Stable-diffusion \cite{rombach22cvpr} & 70.4 & 57.9 & 76.8 & 49.3 & 63.6\\

\end{tabular}
\end{center}
\end{table}

\subsection{Mixing Real and Synthetic Negative Data}
\label{sec:syn_neg_source}

\begin{figure}[ht]
    \centering
    \includegraphics[width=\linewidth]{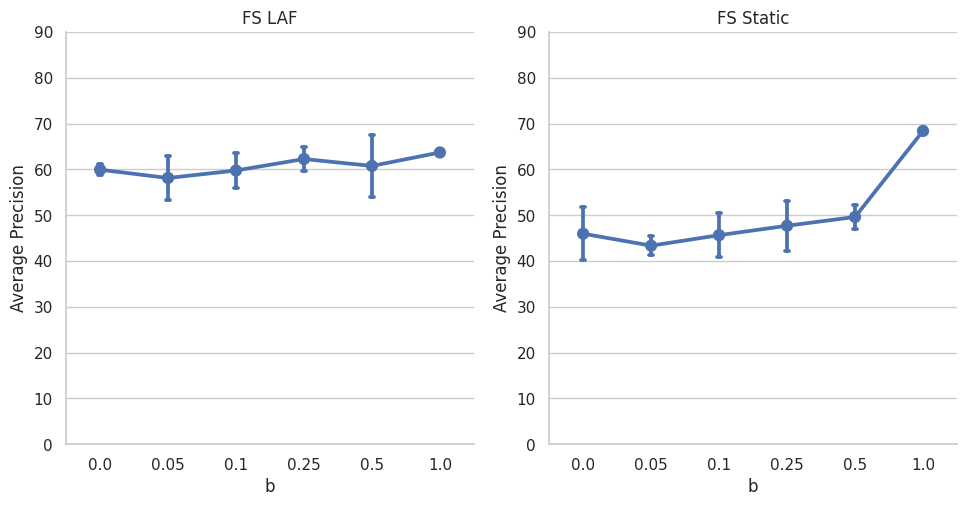}
    \caption{
    Performance of our hybrid anomaly detector when training 
on mixtures of $b$ real and $1-b$ synthetic negatives for $b \in [0,1]$.}
    \label{fig:mix_neg}
\end{figure}

Figure \ref{fig:mix_neg} shows anomaly detection performance when mixing real negatives from ADE20k and synthetic negatives generated by our normalizing flow.
The negative data is mixed according to the hyperparameter $b$ as described in Sec.\ \ref{sec:oss_train_syn}.
We observe varying performance for different values of $b$.
Still, the best performance is achieved when we train solely on real negatives ($b$=1).
Investigating more advanced procedures for mixing real and synthetic negative data is an interesting direction for future work.

\section{Conclusion}

The proposed DenseHybrid  score strives for synergy 
between generative and discriminative anomaly detection
by fusing dataset posterior 
with unnormalized data likelihood.
We eschew the evaluation of 
the intractable normalization constant
by leveraging negative training data.
The negative data can be 
sourced from a general-purpose dataset,
generated by a jointly trained normalizing flow, 
or sampled as a mixture of both sources.
DenseHybrid can be attached
to a desired closed-set model
 in order to attain cutting-edge open-set competence.
 The presented experiments confirm that 
discriminative and generative anomaly detection 
assume different failure modes. 
Furthermore, we observe competitive performance 
on the standard benchmarks
for dense anomaly detection and 
open-set segmentation with negligible computational overhead.
We also propose open-IoU, 
a novel metric for evaluating 
open-set segmentation
and quantifying the performance gap 
between closed-set and open-set setups.
Suitable directions for future work include extensions 
towards open-set panoptics, integration with semantic segmentation approaches
based on class prototypes 
and mask-level recognition, 
 as well as 
further reduction of the gap
between closed-set and open-set performance.

\ifCLASSOPTIONcompsoc
  \section*{Acknowledgments}
\else
  \section*{Acknowledgment}
\fi
This work has been supported by Croatian Science Foundation grant IP-2020-02-5851 ADEPT, NVIDIA Academic Hardware Program, European Regional Development Fund (KK.01.1.1.01.0009 DATACROSS and KK.01.2.1.02.0119 A-Unit), and NextGenerationEU (Croatian NRRP C1.4 R5-I2).
\ifCLASSOPTIONcaptionsoff
  \newpage
\fi



\bibliographystyle{IEEEtran}
%
\vskip -1.5cm
\bibliography{main}



%
\vskip -1.5cm
\begin{IEEEbiography}[{\includegraphics[width=1in,clip,keepaspectratio]{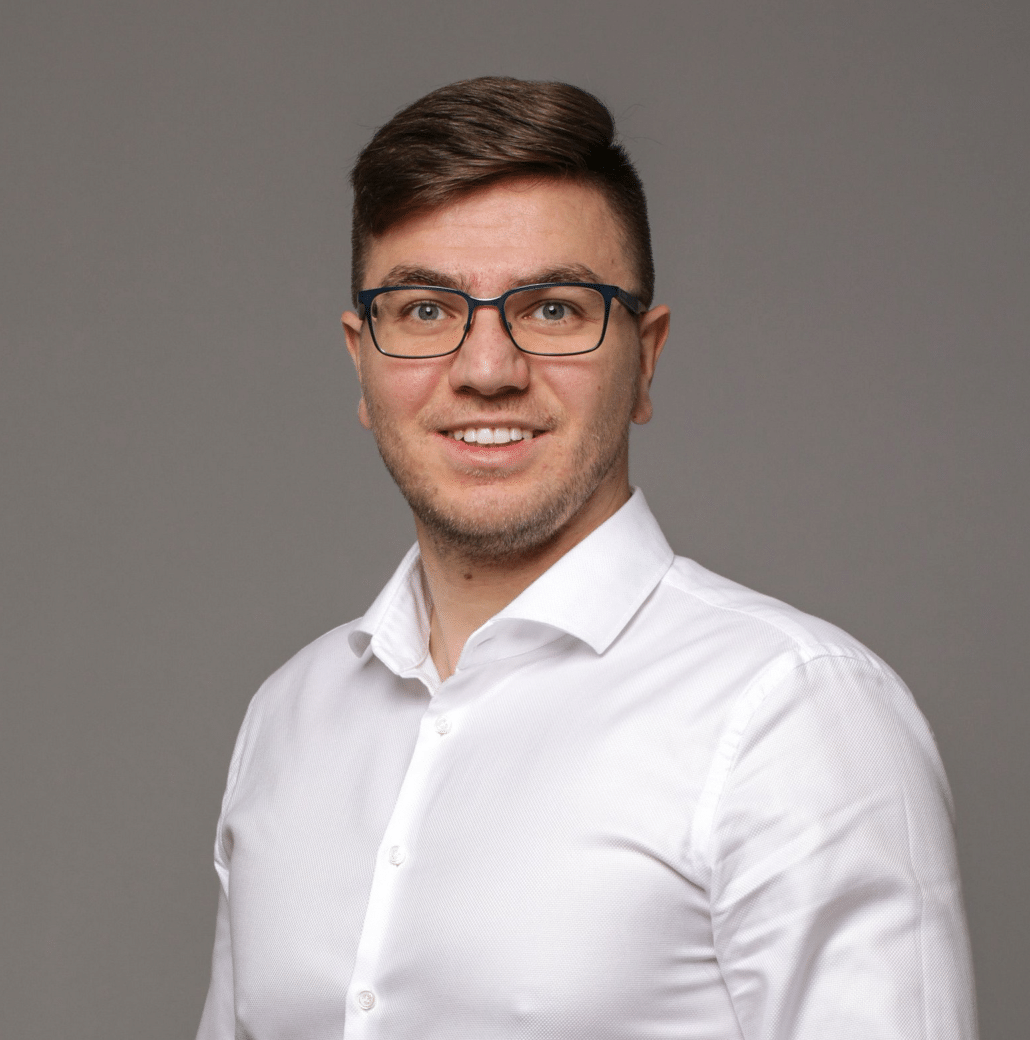}}]{Matej Grcić}
received a M.Sc. degree from the Faculty of Electrical Engineering and Computing in Zagreb. He finished the master study program in
Computer Science in 2020. He is pursuing his Ph.D. degree at University of Zagreb Faculty of Electrical Engineering and Computing.
His research interests include machine learning applications such as recognition in the open world.
\end{IEEEbiography}
\vskip -1.5cm
\begin{IEEEbiography}[{\includegraphics[width=1in,clip,keepaspectratio]{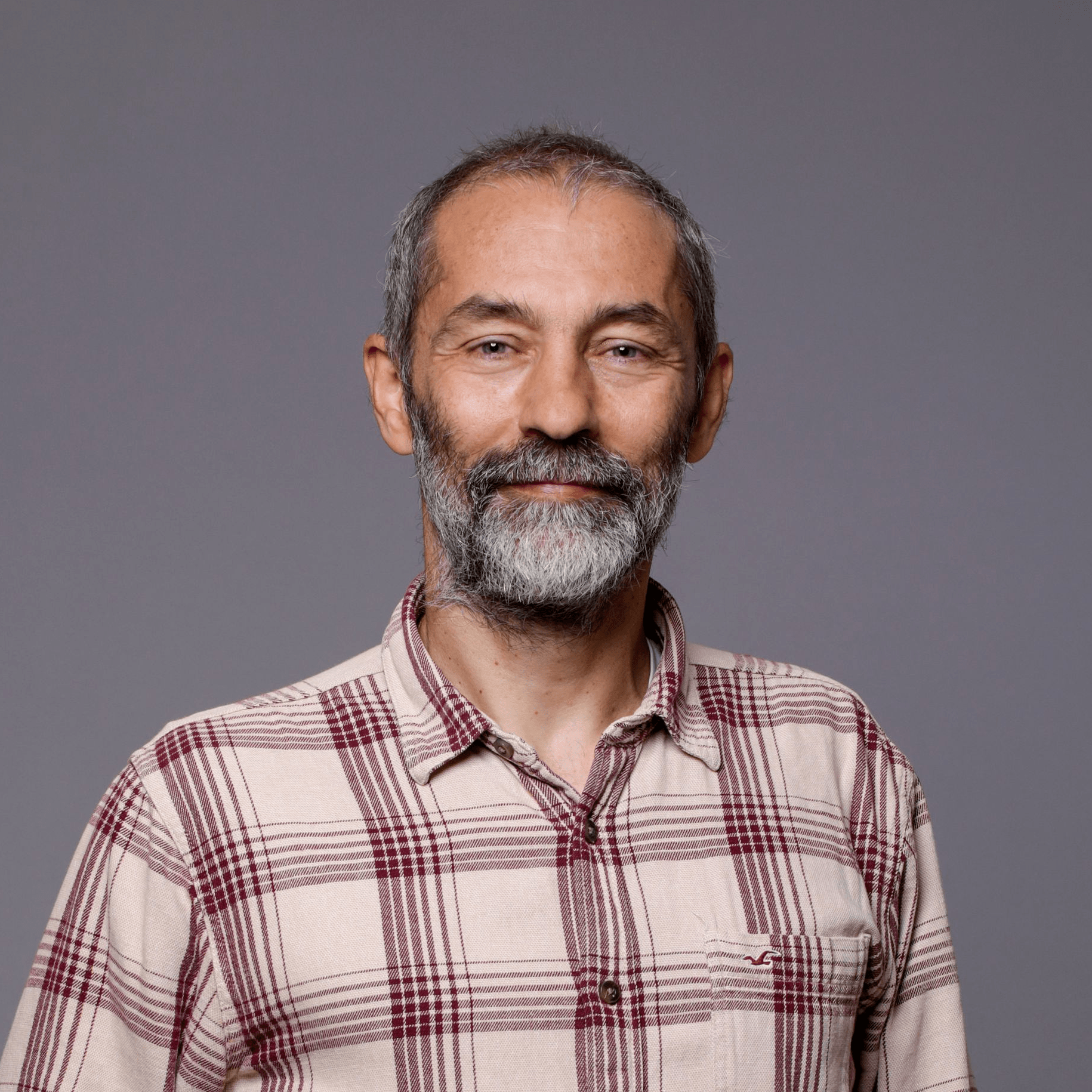}}]{Siniša Šegvić} 
received a Ph.D. degree in computer science from the
University of Zagreb, Croatia. He was a post-doctoral researcher at IRISA
Rennes and at TU Graz. He is currently a full professor at Uni-ZG
FER. His research interests focus on deep models for
classification and dense prediction in unconstrained environments.
\end{IEEEbiography}




\newpage
\section*{Hybrid Open-set Segmentation \\ with Synthetic Negative Data - Appendix}

\subsection*{Limitations}

Our open-set segmentation performance directly depends on the accuracy of the anomaly detector.
While our hybrid anomaly detector surpasses contemporary approaches, it still does not yield perfect performance. Consequently, we observe a performance drop between open-set and closed-set setups (cf.\ Table 3.).
The accuracy of anomaly detectors is also affected by the difficulty of the considered test set.
Anomalies that are semantically similar to the inlier classes \cite{fort21neurips} pose a challenge for contemporary anomaly detectors.
Consequently, the resulting open-set segmentation performance is also affected (cf.\ Table 4.).
Moreover, it may seem that our method can generate samples due to likelihood evaluation being a standard feature of generative models (except GANs). However, sampling unnormalized distributions requires MCMC iteration which is not easily combined with large resolutions and dense loss, at least not with known techniques. Still, our hybrid open-set model delivers competitive performance even without the ability to generate samples.
Finally, if discriminative and generative anomaly detectors are perfectly correlated, our hybrid anomaly detector will not introduce improvements. 
However, this does not happen to be the case in  considered setups
due to different kinds of inductive bias
in generative and discriminative models.

\subsection*{Extended Derivations}
\noindent
\textbf{Loss for data likelihood.}
We present a step-by-step derivation of Eq. (\ref{eq:x_d_0}) as follows.
Note that normalization constant Z cancels out, while LSE denotes log-sum-exp.
\noindent
\begin{align}
L_{\mathbf{x}}(\theta) &= \mathbb{E}_{\mathbf{x} \in D_\mathrm{in}}[- \ln p(\mathbf{x})] - \mathbb{E}_{\mathbf{x} \in D_\mathrm{out}}[-\ln p(\mathbf{x})] \nonumber \\
&= - \mathbb{E}_{\mathbf{x} \in D_\mathrm{in}}[\ln \hat{p}(\mathbf{x}) - \ln{Z}] \nonumber \\
&- \mathbb{E}_{\mathbf{x} \in D_\mathrm{out}}[ - \ln \hat{p}(\mathbf{x}) - \ln{Z}] \nonumber \\
&= - \mathbb{E}_{\mathbf{x} \in D_\mathrm{in}}[\ln \hat{p}(\mathbf{x})] - \ln{Z} + \mathbb{E}_{\mathbf{x} \in D_\mathrm{out}}[\ln \hat{p}(\mathbf{x})] + \ln{Z} \nonumber \\
&= - \mathbb{E}_{\mathbf{x} \in D_\mathrm{in}}[\ln \hat{p}(\mathbf{x})] + \mathbb{E}_{\mathbf{x} \in D_\mathrm{out}}[\ln \hat{p}(\mathbf{x})]  \nonumber \\
&= - \, \mathbb{E}_{D_\mathrm{in}}[\underset{i}{\text{LSE}}(\mathbf{s}_i)]  + \, \mathbb{E}_{D_\mathrm{out}}[\underset{i}{\text{LSE}}(\mathbf{s}_i)]
\end{align}

\noindent
\textbf{Compound loss.} We present a step-by-step derivation of Eq.\ (\ref{eq:final_loss}) as follows.
Recall that the standard cross entropy $L_\text{cls}$ equals to:
\begin{equation}
    L_{\mathrm{cls}}(\theta) = -\, \mathbb{E}_{\mathbf{x}, y \in D_\mathrm{in}}[ \mathbf{s}_{y}] + \mathbb{E}_{\mathbf{x}, y \in D_\mathrm{in}}[\underset{y'}{\text{LSE}}(\mathbf{s}_y')]
\end{equation}
Moreover, the upper bound to data likelihood $L_\textbf{cls}^\text{UB}$ equals to:
\begin{equation}
    L_{\mathbf{x}}^{\mathrm{UB}}(\theta) = - \, \mathbb{E}_{\mathbf{x}, y \in D_\mathrm{in}}[\mathbf{s}_y]  + \mathbb{E}_{\mathbf{x} \in D_\mathrm{out}}[\underset{i}{\text{LSE}}(\mathbf{s}_i)].
\end{equation}
These two losses have a term in common. Consequently, we can omit one of them in the joint loss:
\begin{multline}
     L_{\mathrm{cls}}(\theta) 
 + L_{\mathbf{x}}^{\mathrm{UB}}(\theta) = - \, \mathbb{E}_{\mathbf{x}, y \in D_\mathrm{in}}[\mathbf{s}_y] \\ + \mathbb{E}_{\mathbf{x}, y \in D_\mathrm{in}}[\underset{y'}{\text{LSE}}(\mathbf{s}_y')] + \mathbb{E}_{\mathbf{x} \in D_\mathrm{out}}[\underset{i}{\text{LSE}}(\mathbf{s}_i)].
\end{multline}
The above expression equals to:
\begin{equation}
    L_{\mathrm{cls}}(\theta) 
 + L_{\mathbf{x}}^{\mathrm{UB}}(\theta) = - \mathbb{E}_{\mathbf{x},y \in D_\mathrm{in}}[\ln P(y|\mathbf{x})] - \mathbb{E}_{\mathbf{x} \in D_\mathrm{out}}[\ln \hat{p}(\mathbf{x})] 
\end{equation}
By further adding the data posterior loss and grouping terms according to the expectations we obtain:
\begin{align}
    L(\theta, \gamma) &= L_\text{cls}(\theta) + L_\mathbf{x}^\text{UB}(\theta) + L_\mathbf{d}(\theta, \gamma) \nonumber \\
    &= - \mathbb{E}_{\mathbf{x},y \in D_\mathrm{in}}[\ln P(y|\mathbf{x}) + \ln P(d_\mathrm{in}|\mathbf{x})] \nonumber \\
    & - \mathbb{E}_{\mathbf{x} \in D_\mathrm{out}}[ \ln(1 -  P(d_\mathrm{in}|\mathbf{x})) - \ln \hat{p}(\mathbf{x})].
    \label{eq:ls_fin}
\end{align}
In practice, we introduce loss modulation hyperparameters which control the impact of each loss term (cf. Implementation details in this appendix).

\subsection*{On Effectiveness of Hybrid Anomaly Detector}

Let us consider anomaly scoring function $s: \mathcal{X} \rightarrow \mathbb{R}$ which assigns higher values to anomalies and lower values to normal data.
Without loss of generality, we can assume that the assigned scores are standardized (they have zero mean and unit variance)  since ranking functions are invariant to scaling with positive values and shifting, which are required for standardization.
Let $f: \mathcal{X} \rightarrow \{-1, +1\}$ be a labeling function which outputs +1 if a given input is an anomaly and -1 otherwise.
We can decompose the anomaly scoring function $s$ into a correct labeling $f$ and an error function $\epsilon$:
\begin{equation}
    s(\mathbf{x}) = f(\mathbf{x}) + \epsilon(\mathbf{x}).
\end{equation}
We can compute the expected squared error of a scoring function $s$ as:
\begin{equation}
    \mathcal{E}(s) = \mathbb{E}_\mathbf{x} [(s(\mathbf{x}) - f(\mathbf{x}))^2] =  \mathbb{E}_\mathbf{x} [(\epsilon(\mathbf{x}))^2].
\end{equation}
Note that $\mathcal{E}(s) = 0$ implies perfect separation between inliers and outliers, and therefore leads to perfect score in terms of AP, AUROC and FPR at $\text{TPR}_{95}$.

Our goal is now to show conditions under which the expected square error of hybrid anomaly detector is lower than of the expected error of the best component:
\begin{equation}
    \mathcal{E}(s_H) < \inf\{ \mathcal{E}(s_G), \mathcal{E}(s_D) \}.
    \label{eq:hybrid_ineq}
\end{equation}
The anomaly score $s_G$ is a function of data likelihood and therefore generative anomaly detector.
The anomaly score $s_D$ is a function of the dataset posterior, that is discriminative anomaly detector.
We can define a hybrid anomaly score $s_H$ as an average of the two components:
\begin{equation}
    s_H(\mathbf{x}) := \frac{1}{2} s_D(\mathbf{x}) + \frac{1}{2} s_G(\mathbf{x}).
    \label{eq:therory_hyb_form}
\end{equation}

Then, the expected squared error of the hybrid anomaly score $s_H$ equals:
\begin{align}
     \mathcal{E}(s_H) &= \mathbb{E}_\mathbf{x} [(s_H(\mathbf{x}) - f(\mathbf{x}))^2] \nonumber \\
    &=  \mathbb{E}_\mathbf{x} \left[\left(\frac{1}{2} \epsilon_D(\mathbf{x}) + \frac{1}{2} \epsilon_G(\mathbf{x}) + f(\mathbf{x}) - f(\mathbf{x})\right)^2\right] \nonumber \\
    &=  \mathbb{E}_\mathbf{x} \left[\left(\frac{1}{2} \epsilon_D(\mathbf{x}) + \frac{1}{2} \epsilon_G(\mathbf{x})\right)^2\right] \nonumber \\
    &= \frac{1}{4} \mathbb{E}_\mathbf{x} \left[\left( \epsilon_D(\mathbf{x})\right)^2 + \left(\epsilon_G(\mathbf{x})\right)^2 \right] + \frac{1}{2} \mathbb{E}_\mathbf{x} \left[ \epsilon_D(\mathbf{x})  \epsilon_G(\mathbf{x}) \right] \nonumber \\
    &= \nonumber \frac{1}{4} \mathcal{E}(s_D) + \frac{1}{4} \mathcal{E}(s_G) + \frac{1}{2} \text{cov}(\epsilon_D, \epsilon_G) \\ &+  \frac{1}{2} \mathbb{E}_\mathbf{x} \left[ \epsilon_D(\mathbf{x}) \right] \cdot \mathbb{E}_\mathbf{x} \left[ \epsilon_G(\mathbf{x}) \right] \nonumber \\
    &= \frac{1}{4} \mathcal{E}(s_D) + \frac{1}{4} \mathcal{E}(s_G) + C_1 \rho(\epsilon_D, \epsilon_G) + C_2
    \label{eq:hybrid_error}
\end{align}
Note that $\rho$ represents Pearson's correlation coefficient, while $C_2 = \frac{1}{2} \cdot \mathbb{E}_\mathbf{x}[\epsilon_D(\mathbf{x})] \cdot \mathbb{E}_\mathbf{x}[\epsilon_G(\mathbf{x})]$ and $C_1 = \frac{1}{2} \cdot \text{std}(\epsilon_D) \cdot \text{std}(\epsilon_G)$.
Therefore, by joining (\ref{eq:hybrid_error}) and (\ref{eq:hybrid_ineq}) our goal becomes equivalent to the following inequality:
\begin{equation}
    \frac{1}{4} \mathcal{E}(s_D) + \frac{1}{4} \mathcal{E}(s_G) + C_1 \rho(\epsilon_D, \epsilon_G) + C_2 < \min\{ \mathcal{E}(s_D), \mathcal{E}(s_G) \}.
    \label{eq:bound_1}
\end{equation}
Without loss of generality, we can assume $\mathcal{E}(s_G) < \mathcal{E}(s_D)$. Then, we denote $\mathcal{E}(s_G) = e$ and 
$\mathcal{E}(s_D) = \alpha \cdot e, \alpha > 1$.
We can now rewrite (\ref{eq:bound_1}) as:
\begin{equation}
    \frac{\alpha - 3}{4} e + C_1 \rho(\epsilon_D, \epsilon_G) + C_2 < 0 \; .
    \label{eq:hybrid_effectivness_test}
\end{equation}
We can see that the effectiveness of our hybrid anomaly detector depends on the ratio between the errors of the two components $\alpha = \frac{\max\{ \mathcal{E}(s_G), \mathcal{E}(s_D) \}}{\min\{ \mathcal{E}(s_G), \mathcal{E}(s_D) \}}$, their correlation $\rho$, the level of error $e = \min\{ \mathcal{E}(s_G), \mathcal{E}(s_D) \}$, and constants $C_1, C_2$.
Consequently, equation (\ref{eq:hybrid_effectivness_test}) provides a sufficient condition that the hybrid anomaly detector must satisfy to be effective.
Our hybrid anomaly detector indeed satisfies these conditions in a practical setting  (cf.\ Sec.\ \ref{sec:ablations}).

Finally, we have to show that our hybrid anomaly detector
can be viewed as an ensemble over $s_G$ and $s_D$:
\begin{align}
s_H(\mathbf{x}) &= \frac{1}{2} s_D(\mathbf{x}) + \frac{1}{2} s_G(\mathbf{x}) \\
 &= \frac{1}{2} \frac{s_D'(\mathbf{x}) - \mu_D}{\sigma_D} + \frac{1}{2} \frac{s_G'(\mathbf{x}) - \mu_G}{\sigma_G} \\
     &= \frac{1}{2\sigma_D} s_D'(\mathbf{x})  + \frac{1}{2\sigma_G} s_G'(\mathbf{x}) - \left( \frac{\mu_D}{2\sigma_D} + \frac{\mu_G}{2\sigma_G} \right) \\
     &= \frac{1}{2\sigma} (s_D'(\mathbf{x})  + s_G'(\mathbf{x})) - \left( \frac{\mu_D}{2\sigma_D} + \frac{\mu_G}{2\sigma_G} \right) \\
    &= A \cdot ( \underbrace{(\ln P(d_\mathrm{out}|\mathbf{x}))}_{s_D'(\mathbf{x})} + \underbrace{(- \ln p(\mathbf{x}))}_{s_G'(\mathbf{x})}) + C \\
    &= A \cdot ( \underbrace{(\ln P(d_\mathrm{out}|\mathbf{x}))}_{s_D'(\mathbf{x})} + \underbrace{(- \ln \hat{p}(\mathbf{x})) + \ln Z}_{s_G'(\mathbf{x})}) + C \\
    & \cong \ln P(d_\mathrm{out}|\mathbf{x}) - \ln \hat{p}(\mathbf{x}).
\end{align}
Note that we have assumed $  \sigma = \sigma_D = \sigma_G$.

\subsection*{2D Toy Example Details}

We generate inlier datapoints by sampling the Gaussian mixture $p_\text{in}(\mathbf{x}) = 0.5 \cdot \mathcal{N}(\mu_1, \Sigma_1) + 0.5 \cdot \mathcal{N}(\mu_2, \Sigma_2)$, where $\mu_1 = \mu_2 = 0$ while $\Sigma_1 = \begin{bmatrix}
  0.9 & 0\\ 
  0 & 0.1
\end{bmatrix}$ and $\Sigma_2 = \begin{bmatrix}
  0.071 & 0.071\\ 
  -0.639 & 0.639
\end{bmatrix}$.
The majority of negative training data is located in the first and fourth quadrants of the considered space in order to imitate a finite negative dataset. 
Outlier test data encompass the inlier distribution.
The discriminative anomaly detector is a binary classifier which consists of 4 MLP layers  and ReLU activations.
The generative anomaly detector is an energy-based model with a similar architecture as the binary classifier.
The hybrid anomaly score combines the generative and the discriminative scores as proposed in our method.
We visualize all three anomaly scores on the same scale.
This can be done since the induced rankings
are invariant to monotonic transformations. 
To ensure reproducibility, all samples are generated with the random seed set to 7.
Different seeds also yield similar results.
Source code for this experiment will be available 
at the official DenseHybrid repository \cite{grcic22code}.

\subsection*{Implementation Details}
We construct our open-set models by starting from
any closed-set semantic segmentation model
that trains with pixel-level cross-entropy loss.
We implement the dataset posterior branch $g_\gamma$
as a trainable BN-ReLU-Conv1x1 module.
We obtain unnormalized likelihood as the sum of exponentiated logits.
We fine-tune the resulting open-set models on mixed-content images with pasted negative ADE20k instances (cf.\ Sec.\ \ref{sec:oss_train_aux}) or synthetic negative patches (cf.\ Sec.\ \ref{sec:oss_train_syn}).
In the case of SMIYC, we train LDN-121 \cite{kreso21tits} for 50 epochs in closed-set setup on images from Cityscapes \cite{cordts16cvpr}, Vistas \cite{neuhold17iccv} and Wilddash2 \cite{zendel18eccv} and fine tune for 10 epochs.
In the case of Fishyscapes, we use DeepLabV3+ with WideResNet38 pretrained on Cityscapes \cite{zhu19cvpr}.
We fine-tune the model for 10 epochs on Cityscapes.
In the case of StreetHazards, we train LDN-121 for 120 epochs in the closed-world setting and then fine-tune the open-set model on mixed-content images.
In the case of Pascal-COCO setup we train Segmenter with ViT-B/16 for 100 epochs on inlier data and then fine-tune the model for 10 epochs on real negatives.
During training, we set all background VOC pixels to the mean pixel value. 
This prevents leakage of anomalous semantic content to the inlier representations and hence ensures unbiased performance estimates.
In the case of COCO20/80 we train Segmenter with ViT-B/16 for 80 epochs on inlier data and then fine-tune the model for 9 epochs on real negatives.
In the case of synthetic negative data, we reduce the number of fine-tuning epochs to 5 to prevent overfitting.
We optimize the loss (\ref{eq:ls_fin}) with the following hyperparameters:
\begin{align}
    L(\theta, \gamma) 
    &= - \mathbb{E}_{\mathbf{x},y \in D_\mathrm{in}}[\beta_1 \cdot \ln P(y|\mathbf{x}) + \beta_2 \cdot \ln P(d_\mathrm{in}|\mathbf{x})] \nonumber \\
    & - \mathbb{E}_{\mathbf{x} \in D_\mathrm{out}}[  \beta_3 \cdot \ln(1 -  P(d_\mathrm{in}|\mathbf{x})) - \beta_4 \cdot \ln \hat{p}(\mathbf{x})].
\end{align}
$\beta_1$ always equals 1. For traffic experiments with LDN-121 $\beta_2=\beta_3=0.3$ and $\beta_4=0.03$.
For DLV3+ on traffic scenes $\beta_2=\beta_3=0.1$ and $\beta_4=0.01$ except for Tbl.\ \ref{table:opengan_cs} where $\beta_2=\beta_3=1.5$ and $\beta_4=0.15$.
In the case of Pascal-COCO setup, $\beta_1=1$, $\beta_2=\beta_3 = 1.5$, and $\beta_4=0.15$.
Hyperparameter $\lambda$ from (\ref{eq:flow_total}) always equals 0.03.
Configurations that do not rely on real negative data leverage synthetic data of varying resolutions as generated by DenseFlow-45-6 \cite{grcic21neurips}.
All such experiments pre-train DenseFlow 
with the standard MLE loss on $64\times 64$ crops from road-driving images (except for Pascal-COCO where we pre-train the flow on Pascal images) prior to joint learning.
Our joint fine-tuning experiments last less than 24 hours on a single RTX A5000 GPU.

Open-set evaluation on StreetHazards is conducted as follows.
We partition the  test subset into two folds which correspond to the two test cities - t5 and t6.
We set the anomaly score threshold in order to obtain 95\% TPR on t5, and measure open-mIoU on t6.
Subsequently, we switch the folds and measure open-mIoU on t5.
We compute the overall open-mIoU by weighting these two measurements according to the number of images in the two folds.

We implemented \cite{liu20neurips,sun21neurips,hendrycks19iclr} into our code base by following publicly available implementations.
For the energy fine-tuning \cite{liu20neurips}, we found that the optimal hyperparameters for dense setup are $m_{in}=-15$ and $m_{out}=-5$.
ReAct \cite{sun21neurips} delivers the best results when the method-specific hyperparameter c = 0.99.

\subsection*{Impact of the Segmentation Model}
\label{sec:various_segm}

Table \ref{tbl:models} analyzes the impact of different segmentation models upon which we build our open-set approach.
The top section shows the standard fully convolutional dense classifier DeepLabV3+ \cite{zhu19cvpr} and a convolutional model with near real-time inference LDN-121 \cite{kreso21tits}.
We adopt the public DeepLabV3+ parameterization
trained on Cityscapes train and unlabeled Cityscapes videos,
and train our own LDN-121 on the Cityscapes train.
The bottom section shows transformer-based architectures Segmenter \cite{strudel21iccv} and SWIN + UpperNet \cite{liu21iccv}.
We see that the final open-set performance indeed depends on the segmentation model.
While the convolutional DeepLabV3+ model achieves the best results on FS LAF, the attention-based Segmenter achieves the best results on FS Static. 
These experiments suggest that DenseHybrid can work well with both convolutional and attention-based models.
Following other methods \cite{chan21iccv,vojir21iccv,blum21ijcv,biase21cvpr}
we report results using the convolutional DeepLabV3+ model in most of our main experiments. 
\begin{table}[ht]
\caption{
Impact of different segmentation models on DenseHybrid.
}
\label{tbl:models}
\setlength{\tabcolsep}{5pt}
\begin{center}
\begin{tabular}{llcccc}
\multirow{2}{*}{Backbone} & \multirow{2}{*}{Upsample} & \multicolumn{2}{c}{FS LAF} & \multicolumn{2}{c}{FS Static} \\
 &  & AP &  $\text{FPR}_{95}$ & AP &  $\text{FPR}_{95}$ \\ \hline
WResNet38 & DeepLabV3+ \cite{zhu19cvpr} &  60.5 & 6.0 & 63.1 & 4.2 \\
DenseNet-121 & Ladder \cite{kreso21tits} &  20.9  & 46.2 & 85.8 & 4.1 \\ [.5em]
SWIN-B & UpperNet \cite{liu21iccv} & 17.7  & 59.3 & 91.3 & 1.6 \\
ViT-L/16 & Segmenter \cite{strudel21iccv} & 38.3 & 24.4 & 93.6 & 0.5 \\
\end{tabular}
\end{center}
\end{table}

\subsection*{Ablating Normalizing Flow Learning Objective}
We note that arbitrary loss in negative pixels may not be sufficient to produce adequate synthetic negatives. 
In particular, requiring synthetic negatives to stand out from the inliers 
may be easier to overfit than requiring them 
to produce a uniform prediction over 19 classes.
According to our intuition, 
the latter should provide 
a better learning signal than the former.
Table \ref{tab:ablating_flow_loss} experimentally validates our intuition and shows clear performance gains of requiring uniform classification.
\begin{table}[ht]
\centering
\caption{Validation of the learning objective for our normalizing flow. }
\label{tab:ablating_flow_loss}
\begin{tabular}{lcccc}
\multirow{2}{*}{Loss type} & \multicolumn{2}{c}{FS LAF} & \multicolumn{2}{c}{FS Static} \\
 & AP &  $\text{FPR}_{95}$ & AP &  $\text{FPR}_{95}$ \\ \hline
$L_\textrm{mle} + L_\textrm{d} + L_\textrm{x}^\text{UB} $ & 46.1 & 12.4 & 41.8 & 12.1 \\
$L_\textrm{mle} + L_\textrm{JSD}$ & 60.2 & 7.9 & 52.1 & 7.7
\end{tabular}
\end{table}

\begin{figure*}
    \centering
    \includegraphics[width=\linewidth]{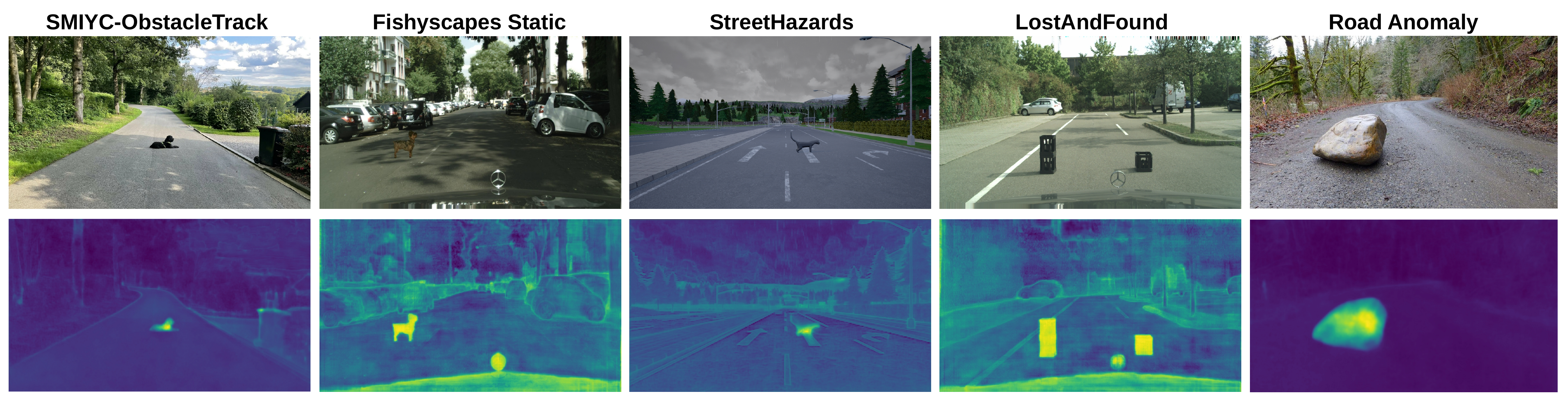}
    \caption{Qualitative performance of the proposed DenseHybrid approach on standard datasets. Top: input images. Bottom: dense maps of the proposed anomaly score. Unknown pixels are assigned with higher anomaly scores as designated in yellow. Such performance of dense anomaly detection enables us to deliver competitive open-set segmentation performance.}
    \label{fig:ood_detector}
\end{figure*}

\subsection*{Impact of the Depth to the Detection Performance}

Road driving scenes typically involve a wide range of depth.
Hence, we explore the anomaly detection performance at different ranges from the camera in order to gain a better insight into the performance of different methods.
We perform these experiments on LostAndFound test \cite{pinggera16iros} since it provides information about the depth in each ground pixel.
Due to errors in the provided disparity maps, we perform our analysis up to 50 meters from the camera.
Table \ref{tbl:distance} indicates that DenseHybrid achieves accurate results even at large distances from the vehicle.
We observe that SynBoost \cite{biase21cvpr} is better than our approach at the shortest range.
However, the computational complexity of image resynthesis precludes real-time deployment of such approaches \cite{biase21cvpr,lis19iccv,xia20eccv} on present hardware as we show next.

\begin{table}[ht]
\centering
\caption{Anomaly detection performance at different distances from camera.
}
\label{tbl:distance}
\begin{tabular}{l|cccccccc}
\multirow{2}{*}{Range} & \multicolumn{2}{c}{MSP \cite{hendrycks17iclr}} & \multicolumn{2}{c}{ML \cite{hendrycks22icml}} & \multicolumn{2}{c}{SynBoost \cite{biase21cvpr}} & \multicolumn{2}{c}{DH (ours)} \\
 & AP & FPR & AP & FPR & AP & FPR & AP & FPR \\ \hline
5-10 & 28.7 & 16.4 & 76.1 & 5.4 & \textbf{93.7} & \textbf{0.2} & 90.7 & 0.3 \\
10-15 & 28.8 & 29.7 & 73.9 & 16.2 & 78.7 & 17.7 & \textbf{89.8} & \textbf{1.1} \\
15-20 & 26.0 & 28.8 & 78.2 & 5.9 & 76.9 & 25.0 & \textbf{92.9} & \textbf{0.6} \\
20-25 & 25.1 & 44.2 & 69.6 & 12.8 & 70.0 & 23.3 & \textbf{89.1} & \textbf{1.4} \\
25-30 & 29.0 & 41.3 & 72.6 & 9.5 & 65.6 & 18.8 & \textbf{89.5} & \textbf{1.4} \\
30-35 & 26.2 & 47.8 & 70.2 & 10.0 & 58.5 & 27.4 & \textbf{87.7} & \textbf{2.5} \\
35-40 & 29.6 & 44.7 & 71.0 & 9.8 & 59.8 & 25.4 & \textbf{85.0} & \textbf{3.7} \\
40-45 & 31.7 & 43.2 & 74.0 & 9.8 & 60.0 & 25.8 & \textbf{85.6} & \textbf{4.7} \\
45-50 & 33.7 & 45.3 & 73.9 & 11.0 & 53.3 & 29.9 & \textbf{82.1} & \textbf{6.3}
\end{tabular}
\end{table}

\subsection*{Inference Speed}
\label{sec:speed}

Table \ref{tbl:speed} compares computational overheads of prominent anomaly detectors on two-megapixel images.
All measurements are averaged over 200 runs on RTX3090.
DenseHybrid involves a negligible computational overhead of 0.1 GFLOPs and 2.8ms.
These experiments indicate that image resynthesis is not applicable for real-time inference on present hardware.
\begin{table}[ht]
\begin{center}
\caption{Computational overhead of prominent anomaly detectors over the baseline semantic segmentation model when inferring on two-megapixel images.
The inference time is in milliseconds.
}
\label{tbl:speed}
\begin{tabular}{lcccc}

Method & \multicolumn{1}{c}{Resynth.} & \multicolumn{1}{c}{Inf. time} & \multicolumn{1}{c}{FPS} & GFLOPs \\\hline 
SynBoost \cite{biase21cvpr} & \cmark & 1055.5 & $<$1 & - \\
SynthCP \cite{xia20eccv} & \cmark & 146.9 & $<$1 & 4551.1 \\
LDN-121 \cite{kreso21tits} & \xmark & 60.9 & 16.4 & 202.3 \\ 
LDN-121 + SML \cite{jung21iccv} & \xmark & 75.4 & 13.3 & 202.6\\
LDN-121 + DH (ours) & \xmark & \textbf{63.7}  & \textbf{15.7} & \textbf{202.4} 
\end{tabular}
\end{center}
\end{table}

\subsection*{Computational Advantage of Translational Equivariance}

Translational equivariance ensures
 efficient inference due to opportunity
 to share latent representations 
 across neighbouring estimates.
 This makes such models
 much more efficient than their
  sliding-window counterparts.
 We illustrate this point by considering 
 to replace our hybrid unnormalized density
 with a non-equivariant image-wide generative model.
 We consider a medium capacity normalizing flow 
 (Glow with 25 coupling layers)
 that observes a channel-wise concatenation 
 of a $128 \times 128$ input crops and 
 suitably upsampled semantic features.
 Note that we do not train the flow
 since that would exceed the scope of this paper
 and would likely represent a contribution on its own.
 Instead, we only  measure the computational strain 
 of applying the considered flow 
 in a sliding window with stride 8.
 We observe that such dense estimates
 of the probability density function
 require $2.6\times$ more MACs and $42 \times$ more wall-clock time for dense open-set inference compared to DenseHybrid pipeline. 
 This confirms our argument from the introduction
 that the computational cost of non-equivariant density 
 would preclude practical applications of generative anomaly detection in dense prediction contexts.

\subsection*{Alternative Synthetic Negatives}
\begin{figure}[h]
    \centering
    \includegraphics{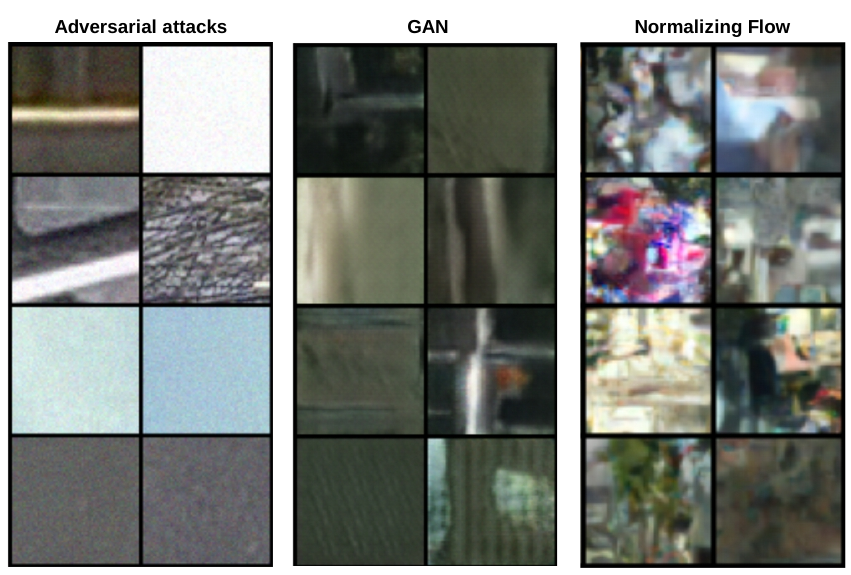}
    \caption{Comparison of synthetic negatives produced by local adversarial attacks, jointly trained GAN, and the proposed jointly trained normalizing flow. Our method produces synthetic negatives 
that diverge from the inliers 
while being more visually diverse
than the other two approaches.}
    \label{fig:train_negatives}
\end{figure}

Figure  \ref{fig:train_negatives} shows synthetic negatives produced by local adversarial attacks \cite{besnier21iccv}, jointly trained GAN \cite{lee18iclr},  and the proposed normalizing flow.
We observe that adversarial attacks appear as blurred patches of inlier scenes.
This may make them an inadequate proxy for test-time anomalies due to requiring a lot of capacity 
to discriminate them from the inliers 
\cite{madry18iclr}.
Similarly, it is well known that GAN samples 
struggle to achieve visual variety \cite{lucas19nips}. 
Contrary, our normalizing flow produces negatives which differ from the inlier scenes and are visually diverse. 
Different from our jointly trained normalizing flow, 
the text-to-image models cannot produce 
dataset-specific samples 
and require textual descriptions.
Consequently, they are not in the same ballpark as our method.
Furthermore,
the design of textual prompts for generating synthetic negative data 
is still an open problem, which requires further
work that is out of the scope of our work.
Note that the quantitative comparison of the three approaches can be found in Table \ref{tbl:syn_source}.

\subsection*{Qualitative Open-set Experiments}
\begin{figure}[h]
    \centering
    \includegraphics[width=\linewidth]{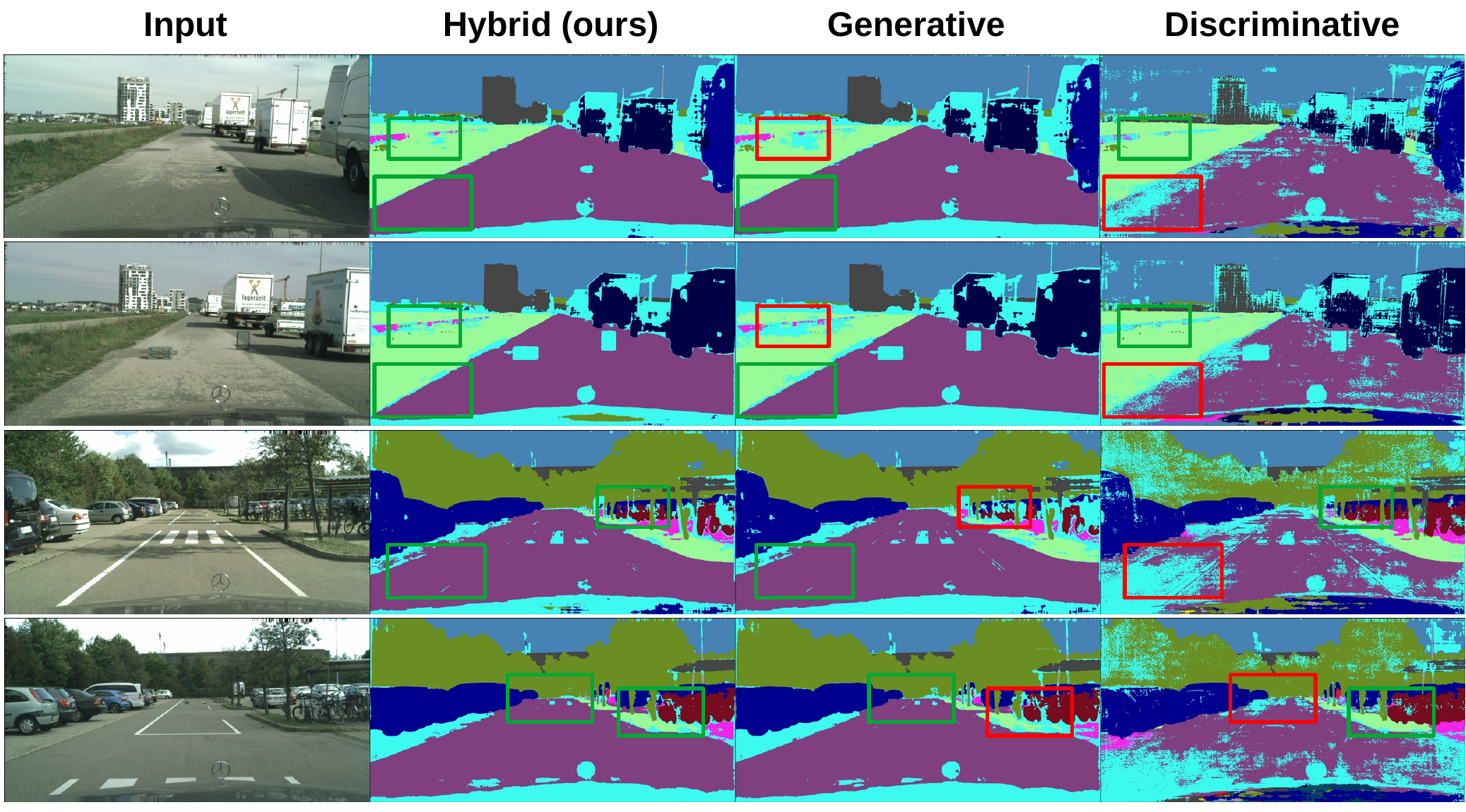}
    \caption{Open-set segmentation on FS LAF with discriminative, generative and hybrid anomaly detection.
    Hybrid anomaly detection yields the lowest FPR95 metric. Red and green boxes indicate the abundance and absence of false positive anomalies, respectively. Please zoom in for a better view.}
    \label{fig:ablate_osr_traffic}
\end{figure}

Figure \ref{fig:ablate_osr_traffic} shows open-set segmentation performance on traffic scenes from Fishyscapes LAF val with different anomaly detectors.
Our hybrid anomaly detector yields the lowest false-positive count for $\text{TPR}=95\%$.

\subsection*{Performance on Fishyscapes val and Road Anomaly}

Table \ref{tab:fs_val} presents performance on Road Anomaly\cite{lis19iccv} and validation subsets of Fishyscapes.
The top section presents methods which do not train on real negative data. 
The bottom section presents methods which train on real negative data.
Our method performs competitively with respect to the previous work in both setups.
\begin{table}[ht]
\centering
 \caption{Validation performance on Road Anomaly and Fishyscapes val.
 }
\label{tab:fs_val}
\setlength{\tabcolsep}{5pt}
\begin{footnotesize}
\begin{tabular}{lcccccc}
\multirow{2}{*}{Model} & \multicolumn{2}{c}{RA} & \multicolumn{2}{c}{FS LAF} & \multicolumn{2}{c}{FS Static} \\
  & AP &  $\text{FPR}_{95}$ & AP &  $\text{FPR}_{95}$ & AP &  $\text{FPR}_{95}$\\ \hline
MSP \cite{hendrycks17iclr} & 15.7 & 71.4 & 4.6 & 40.6 & 19.1 & 24.0\\
ML \cite{hendrycks22icml} & 19.0 & 70.5 & 14.6 & 42.2 & 38.6 & 18.3\\
SML \cite{jung21iccv} & 25.8 & 49.7 & 36.6 & 14.5 & 48.7 & 16.8 \\
SynthCP \cite{xia20eccv} & 24.9 & 64.7 & 6.5 & 46.0 & 23.2 & 34.0 \\
Emb. Density \cite{blum21ijcv} &  - & - & 4.1 & 22.3 & - & - \\
SynDenseHybrid & \textbf{35.1} & \textbf{37.2} & \textbf{60.2} & \textbf{7.9} & \textbf{52.1} & \textbf{7.7} \\[.5em]
SynBoost\cite{biase21cvpr} & 38.2 & 64.8 & 60.6 & 31.0 & \textbf{66.4} & 25.6\\
OOD head \cite{bevandic22ivc} & - & - & 45.7 & 24.0 & - & - \\
Energy \cite{liu20neurips} & 19.5 & 70.2 & 16.1 & 41.8 & 41.7 & 17.8 \\
DenseHybrid & \textbf{63.9} & \textbf{43.2} & \textbf{60.5} & \textbf{6.0} & 63.1 & \textbf{4.2}
\end{tabular}
\end{footnotesize}
\end{table}

\begin{figure}[ht]
    \centering
    \includegraphics[width=\linewidth]{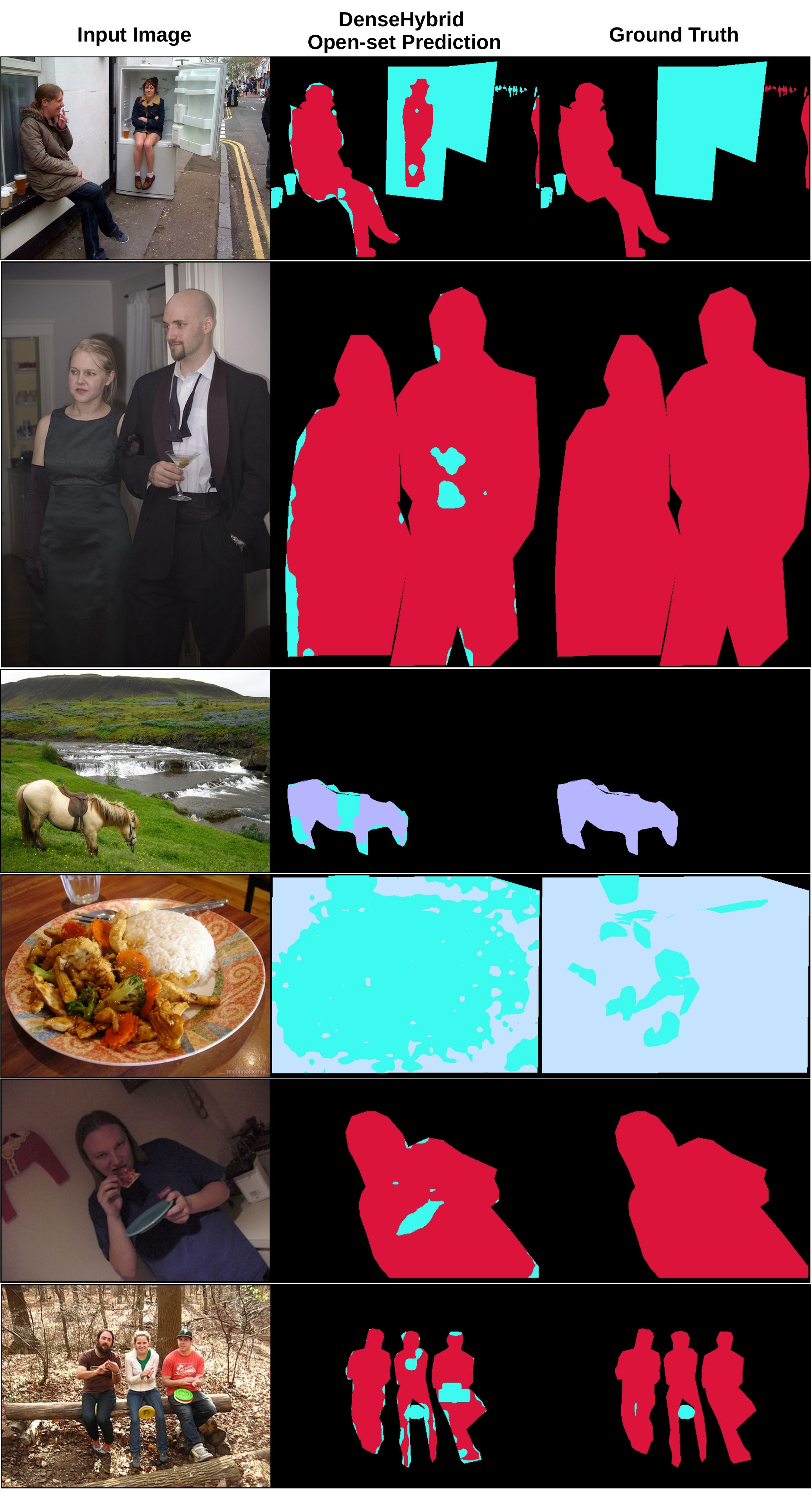}
    \caption{Examples of mistakes committed by DenseHybrid on COCO20/80.
    The mistakes often coincide with labelling errors.
    For visualization purposes, we override predictions in void pixels (dark).
    }
\end{figure}

\subsection*{More Qualitative Examples}

\begin{figure}[ht]
    \centering
    \includegraphics[width=\linewidth]{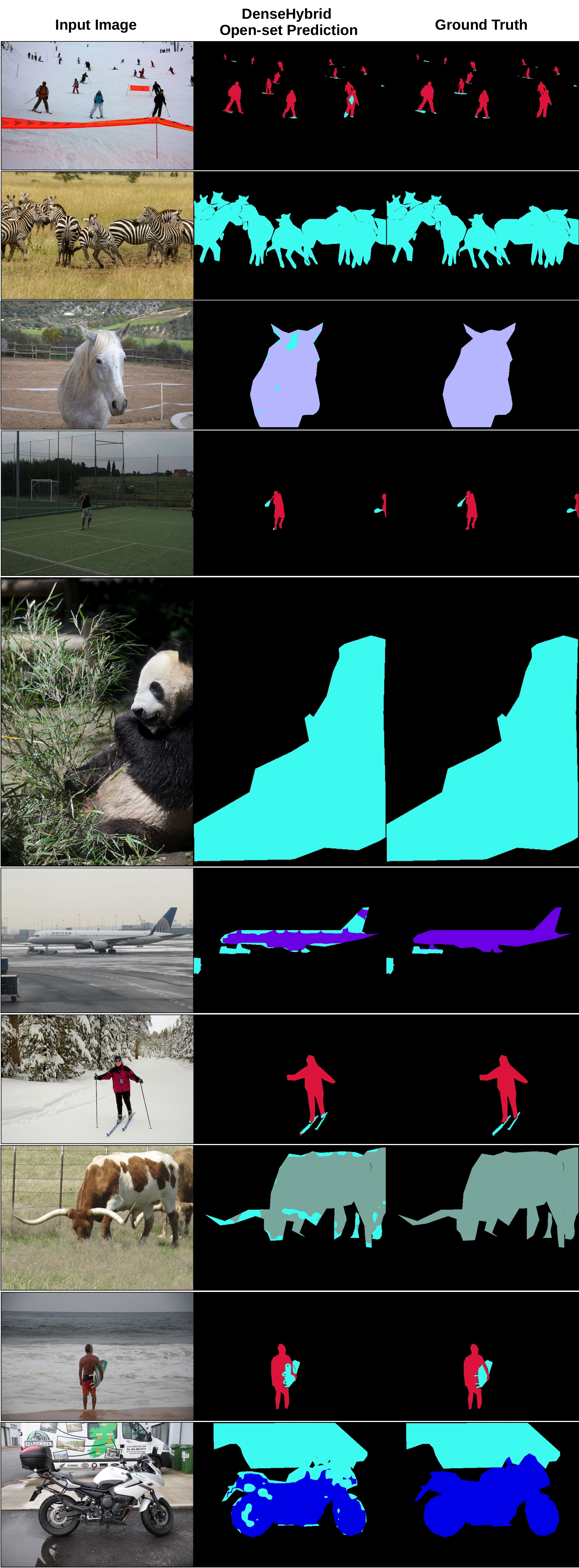}
    \caption{More qualitative examples of open-set segmentation on COCO20/80.
    Unknown unknown pixels are denoted in cyan.
    For visualization purposes, we override predictions in void pixels (dark).
    }
    \label{fig:osr_more}
\end{figure}

Figure \ref{fig:osr_more} shows more qualitative examples for open-set segmentation on the COCO20/80 dataset.
We observe that DenseHybrid can detect both large and small unknowns.

\end{document}